\pdfoutput=1
\documentclass[11pt]{article}

\usepackage[preprint]{acl}

\usepackage{times}
\usepackage{latexsym}
\usepackage[T1]{fontenc}
\usepackage[utf8]{inputenc}
\usepackage{microtype}
\usepackage{graphicx}
\usepackage{booktabs}
\usepackage{amsmath}
\usepackage{amssymb}
\usepackage{multirow}
\usepackage{enumitem}
\usepackage{algorithm}
\usepackage{algpseudocode}
\usepackage{listings}
\usepackage[most]{tcolorbox}
\makeatletter
\g@addto@macro{\UrlBreaks}{\do\-}
\makeatother
\newtcblisting{promptbox}[2][]{%
  enhanced,
  breakable,
  listing only,
  colback=gray!3,
  colframe=black!55,
  coltitle=black,
  colbacktitle=gray!14,
  fonttitle=\bfseries\footnotesize,
  title={#2},
  boxrule=0.4pt,
  arc=1pt,
  left=3pt,
  right=3pt,
  top=3pt,
  bottom=3pt,
  before skip=6pt,
  after skip=8pt,
  listing options={%
    basicstyle=\ttfamily\scriptsize,
    breaklines=true,
    breakatwhitespace=false,
    columns=fullflexible,
    keepspaces=true,
    showstringspaces=false,
    upquote=true,
    literate=
      {∀}{{$\forall$}}1
      {∃}{{$\exists$}}1
      {∧}{{$\land$}}1
      {∨}{{$\lor$}}1
      {¬}{{$\neg$}}1
      {→}{{$\to$}}1
      {↔}{{$\leftrightarrow$}}1
      {⊕}{{$\oplus$}}1
  },
  #1
}

\tcbset{caseinput/.style={after skip=0pt, listing options={basicstyle=\ttfamily\scriptsize,
  breaklines=true, breakatwhitespace=true, columns=fullflexible, keepspaces=true,
  showstringspaces=false, upquote=true, escapeinside={(*}{*)}}}}

\newcommand{\best}[1]{\textbf{#1}}
\newcommand{\snd}[1]{\underline{#1}}

\newcommand{\Dval}{\ensuremath{\mathcal{D}^{+}}}

\title{Prove What You Can: Hybrid Neurosymbolic Reasoning for Logical Natural Language Inference}

\author{Ada Canoğlu \\
  Sabancı University \\
  Istanbul, Türkiye \\
  \texttt{adacanoglu@sabanciuniv.edu} \\\And
  Sadiq Qara \\
  Independent Researcher \\
  Baku, Azerbaijan \\
  \texttt{sadiqqarayev04@gmail.com} \\}

\begin{document}
\maketitle

\begin{abstract}
Logical reasoning over natural language requires both semantic interpretation and deductive search. Large language models (LLMs) can propose plausible intermediate derivations, but they often fail to compose them into a complete proof and overpredict uncertainty. Neurosymbolic methods improve search by translating an entire problem into first-order logic and handing it to a prover, but a single bad clause can make the whole translation non-executable. We present \textbf{Hybrid Reasoning}, a three-stage pipeline that decouples proposing a proof, validating its steps, and deciding a label. An LLM proposes a structured proof tree with a raw verdict; a theorem prover validates each derivation locally, keeping only the subset it can prove; and an \emph{uncertainty-gated repair} attempts to re-derive the label from the premises plus the validated derivations, when the raw verdict is \textsc{Uncertain}. Across four benchmarks (FOLIO, ProofWriter, ProverQA, PrOntoQA) and five contemporary LLMs, Hybrid Reasoning is the top-performing method on the two hardest datasets, improving on the strongest baseline by $5.3$ points on FOLIO and $4.9$ on ProverQA, and reaches ceiling-level accuracy on the easiest. The repair stage operates at $78$--$98\%$ precision, producing $326$ correct relabellings against $19$ regressions.
\end{abstract}

\section{Introduction}
\label{sec:intro}

Logical reasoning over natural language requires both semantic interpretation and deductive search. A system must map linguistically varied premises into the right semantic relations, but it must also determine whether a hypothesis $h$ follows from a set of premises $P$, whether its negation follows, or whether the evidence is insufficient. Large language models are flexible semantic interpreters, yet their multi-step reasoning is often not \emph{faithful}: intermediate steps may not entail their conclusions, and the final answer can be right for the wrong reasons or wrong despite a mostly useful chain \citep{wei2022chain,kojima2022large,PrOntoQA}. For logical natural language inference (NLI), this is especially costly because a single unsound or missing step can flip the label.

Two families of methods address different sides of this problem. \emph{Neurosymbolic} approaches translate the whole problem into a formal language and discharge it with a symbolic solver \citep{pan-etal-2023-logic,olausson2023linc,ye2023satlm}. These methods work when the translation is faithful and executable, but one mistranslated quantifier, missing premise or undeclared predicate makes the whole program fail. \emph{Process-level} approaches instead score or search over individual reasoning steps \citep{lightman2023letsverify,yang2022nlproofs,kamoi2026fover}, but typically rely on learned verifiers that are themselves approximate and offer no symbolic guarantee.

Proof-tree reasoning and formalization fail in different ways. A proof-tree model often derives correct intermediate facts but stops before the hypothesis and answers Uncertain, whereas a monolithic formalizer fails whenever part of the input is mistranslated. We therefore use LLM proof trees as \emph{semantic proposal mechanisms} and theorem provers as \emph{local agreement filters}. Rather than trusting an entire generated proof tree or requiring a complete global formalization, our method validates individual candidate derivations and reuses only those accepted by the symbolic stage.

\begin{figure*}[tp]
  \centering
  \includegraphics[width=0.99\textwidth]{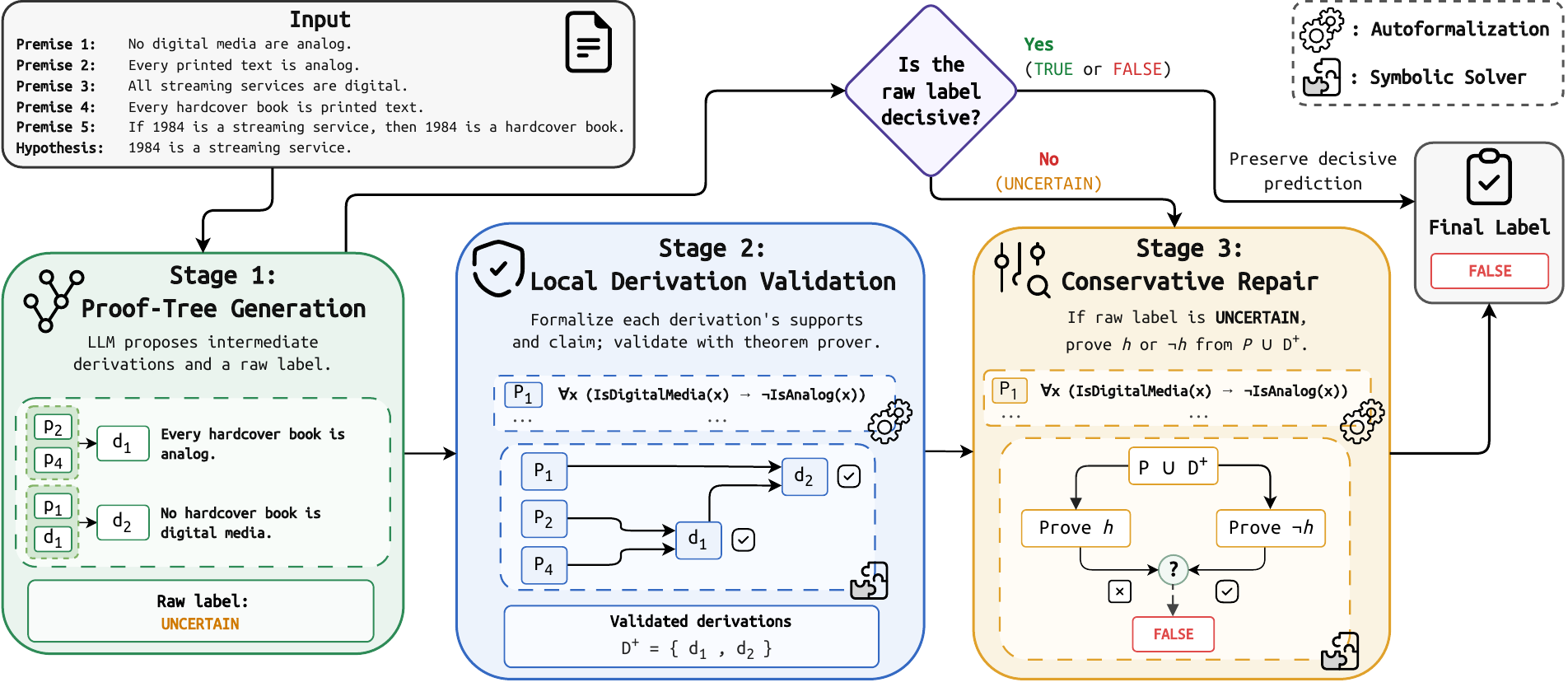}
  \caption{\textbf{Hybrid Reasoning.} \textbf{Stage~1} an LLM proposes derivations $d_i$ with supports $S_i$ and a raw verdict. \textbf{Stage~2} each derivation is validated as a local entailment problem $S_i\vdash d_i$, yielding the validated set $\Dval=\{d_i : S_i\vdash d_i\}$. \textbf{Stage~3} if the raw verdict is decisive (\textsc{True}/\textsc{False}) the prediction is preserved; otherwise \emph{uncertainty-gated repair} attempts to prove $h$ or $\neg h$ from $P\cup\Dval$, repairing the label only if one side is discharged.}
  \label{fig:framework}
\end{figure*}

\paragraph{Hybrid Reasoning.} We introduce a pipeline that factorises these operations (Figure~\ref{fig:framework}). The name covers only this combination of per-derivation validation and uncertainty-gated repair; it is unrelated to LLMs that switch between thinking and non-thinking modes \citep{yang2025qwen3technicalreport}. In \textbf{Stage~1}, an LLM proposes a structured proof tree: a sequence of derivations $d_1,\dots,d_k$, where each $d_i$ ($i\in\{1,\dots,k\}$) cites a support set $S_i\subseteq P\cup\{d_j : j<i\}$, together with a \emph{raw verdict} in $\{\textsc{True},\textsc{False},\textsc{Uncertain}\}$. In \textbf{Stage~2}, each derivation is checked as its own small entailment problem: its supports become local premises and its claim the local conclusion, both autoformalized and discharged by a prover; we keep only the validated subset $\Dval=\{d_i : S_i\vdash d_i\}$. In \textbf{Stage~3}, an \emph{uncertainty-gated repair} fires only when the raw verdict is \textsc{Uncertain}: it attempts to prove $h$, then $\neg h$, from $P\cup\Dval$, and relabels the instance by the first side it discharges, and otherwise leaving it \textsc{Uncertain}. Confident raw \textsc{True}/\textsc{False} verdicts are returned unchanged. The symbolic stage does limited work: it searches and validates within the derivational structure the LLM already proposed.

The autoformalization engine that powers Stage~2 is also a standalone baseline, letting us measure exactly what per-derivation validation and gated repair add on top of monolithic translation, on identical models and splits.

\paragraph{Contributions.}
\begin{enumerate}[noitemsep,topsep=2pt,leftmargin=1.4em]
\item We propose \textbf{Hybrid Reasoning}, a three-stage neurosymbolic pipeline that  separates proposing, validating, and deciding a logical NLI label, and validates reasoning \emph{per derivation} rather than per problem (\S\ref{sec:method}).
\item We introduce \textbf{uncertainty-gated repair}: a conservative relabelling step that trusts only symbolically validated derivations ($\Dval$) and fires only on the model's own \textsc{Uncertain} verdicts (\S\ref{sec:method}).
\item Across four benchmarks and five LLMs, Hybrid Reasoning is strongest on the hardest datasets (FOLIO and ProverQA), ceiling-level on PrOntoQA, and highly competitive on ProofWriter; direct repair is high-precision and net-positive (\S\ref{sec:results}--\S\ref{sec:analysis}).\footnote{Code will be released on GitHub}
\end{enumerate}

\section{Related Work}
\label{sec:related}

\paragraph{Full-translation neurosymbolic reasoning.} Logic-LM \citep{pan-etal-2023-logic} formulates a problem, hands it to one of several solvers, and self-refines on solver error messages; LINC \citep{olausson2023linc} parses each problem into first-order logic and aggregates $k$ prover calls by majority vote; SatLM \citep{ye2023satlm} casts reasoning as satisfiability. All commit to a \emph{single global} formalisation, so one faulty clause can forfeit the whole derivation, and LINC's own analysis attributes a large share of errors to translation faults. Hybrid Reasoning validates derivations one at a time and reuses only those the prover accepts; whole-problem formalization appears here only as a baseline.

\paragraph{Decomposition and compositional translation.} CLOVER \citep{ryu2025divide} translates each sentence compositionally and verifies candidate formulas by logical-equivalence or counter-interpretation; Decompose-and-Formalise \citep{quan2026decompose} builds an entailment tree, verifies it bottom-up, and locally refines failing steps; EPF \citep{lee2025epf} trains translators to preserve entailment. These verify \emph{translations} or \emph{rewrite} explanations until they pass. Hybrid Reasoning instead asks a different question per step (whether the cited supports actually entail this claim) and \emph{discards} failing steps without rewriting them, treating the surviving set as evidence for a gated decision.

\paragraph{Step verification and process supervision.} Process reward models \citep{lightman2023letsverify} and formally supervised variants \citep{kamoi2026fover} learn to score steps; NLProofS \citep{yang2022nlproofs} searches proofs under a learned verifier. We replace the learned verifier with a sound prover invoked per derivation, requiring no step-level supervision; the cost is that derivations which fail to autoformalize are left out of $\Dval$. FANS \citep{fans} also uses a sound prover to improve natural-language reasoning, but at the answer level: it translates candidate answers to math problems into Lean4 and selects among them, whereas we verify intermediate first-order derivations and use the verified subset only to resolve abstentions.

\paragraph{Proof generation and certified decoding.} Earlier work generates natural-language proofs or entailment trees \citep{clark2020transformers,dalvi2021entailmentbank,saha2021multiprover,tafjord2022entailer}, and certified decoding constrains generation to be sound \citep{poesia2023logicguide}. Hybrid Reasoning borrows the proof-tree representation but adds a post-hoc symbolic filter, and works with any black-box LLM API. Symbolic chain-of-thought \citep{xu2024symbcot} and self-consistency \citep{wang2023selfconsistency} check reasoning within the LLM, whereas Hybrid Reasoning uses an external prover.

\paragraph{Benchmarks.} We evaluate on FOLIO \citep{han-etal-2024-folio}, ProofWriter \citep{tafjord2021proofwriter}, ProverQA \citep{qi2025proverqa}, and PrOntoQA \citep{PrOntoQA}; related benchmarks include P-FOLIO \citep{han2024pfolio}, LogicBench \citep{parmar2024logicbench}, Multi-LogiEval \citep{patel2024multilogieval}, and SATBench \citep{wei2025satbench}. General NLI corpora \citep{bowman2015large,williams2018broad,nie2020adversarial} lack the explicit logical structure our symbolic stage requires.

\section{Method}
\label{sec:method}

\subsection{Problem setting and notation}
An instance is a triple $(P,h,\ell)$ where $P=\{p_1,\dots,p_n\}$ is a set of natural-language premises, $h$ is a hypothesis, and $\ell$ is the gold label. The label space is $\{\textsc{True},\textsc{False},\textsc{Uncertain}\}$ under the open-world reading used by FOLIO, ProofWriter, and ProverQA; PrOntoQA is the binary closed-world case. A method maps $(P,h)$ to a predicted label $\hat\ell$. We use three label terms throughout: the \emph{gold} label $\ell$ from the dataset; the \emph{raw} verdict $r$ that the LLM outputs in Stage~1 (reported as \emph{Hybrid Raw}); and the \emph{final} label $\hat\ell$ after Stage~3 (\emph{Hybrid Final}, or simply \emph{Hybrid} in tables).

A derivation is a natural-language claim $d_i$ together with a support set $S_i\subseteq P\cup\{d_j:j<i\}$. We use validation as an operational agreement signal between the proof-tree stage and the symbolic stage. We do not assume that validated derivations constitute faithful explanations of the model's internal reasoning, nor that failed validation implies semantic invalidity: the prover certifies entailment only relative to the generated formalization.

\subsection{Stage 1: Proof-tree generation}
The LLM is prompted to emit a structured object rather than free text, using the provider's JSON-schema or JSON-object output mode where available (Appendix~\ref{app:structured} discusses this choice and its alternatives). The object contains: (i) a \emph{goal analysis} naming a \texttt{true\_goal} and a \texttt{false\_goal}, each decomposed into subgoals and annotated with a \emph{main connective} describing the conclusion's logical shape (a single literal, a conjunction, or a disjunction); (ii) a \emph{proof tree}, an ordered list of derivations $d_1,\dots,d_k$ where each $d_i$ records a claim and a support set $S_i\subseteq P\cup\{d_j : j<i\}$ of premise indices and earlier derivations; and (iii) a \emph{raw verdict} $r\in\{\textsc{True},\textsc{False},\textsc{Uncertain}\}$. We parse the object, check required fields, and structurally validate that every cited support exists and is acyclic; malformed outputs trigger a single repair-prompt that returns the parser error (full prompt schema and one-shot demonstrations in Appendix~\ref{app:prompts}). If the object is still invalid after this retry, the instance receives a chain-of-thought prediction instead; this happened in 2 of 4{,}085 Hybrid runs (\S\ref{app:executability} and \S\ref{app:impl}). The raw verdict $r$ is the prediction of an LLM-only ``Hybrid Raw'' system and is the quantity the later stages may revise.

\subsection{Stage 2: Local derivation validation}
Each derivation is re-cast as a self-contained entailment problem and checked symbolically. For $d_i$, its supports $S_i$ become local premises and its claim becomes the local conclusion. The pair is passed to an autoformalization module (the same module used as our standalone \emph{Autoformalization} baseline), which produces a shared vocabulary and first-order formulas, checks vocabulary consistency, lowers the result to the Prover9 theorem prover \citep{mccune2010prover9}, and runs it. The \emph{validated set} is
\[
\Dval \;=\; \{\, d_i \;:\; S_i \vdash d_i \,\},
\]
i.e.\ exactly the derivations the prover discharges from their cited supports. Target matches that link derivations to the goal subgoals are validated the same way. A derivation that fails to autoformalize or is not provable is \emph{omitted} from $\Dval$; it is never treated as evidence against the model. The per-instance proportion $|\Dval|/k$ yields a \emph{validation status}: \emph{fully}, \emph{partially}, or \emph{not} validated (plus \emph{empty tree}), which we later show predicts reliability.

\subsection{Stage 3: Uncertainty-gated repair}
\label{sec:repair}
If the raw verdict $r$ is decisive (\textsc{True} or \textsc{False}), Hybrid Reasoning preserves it: the validated derivations are not allowed to overturn a confident model prediction (Appendix~\ref{app:ablations} ablates this gate). Repair fires only when $r=\textsc{Uncertain}$. In that case we assemble the repair premise set $P\cup\Dval$ (the original premises augmented with only the proved derivations) and attempt to discharge a hierarchy of conclusions on each side: the original conclusion, then the side goal, then its subgoals, combined according to the main connective (a single literal requires its lone subgoal; a disjunctive goal accepts any subgoal; a conjunctive goal requires all). If the true side is discharged we relabel \textsc{True}; if the false side is discharged, \textsc{False}; if neither, the instance remains \textsc{Uncertain}. Algorithm~\ref{alg:hybrid} summarises the procedure.

\begin{algorithm}[htp]
\small
\caption{Hybrid Reasoning}
\label{alg:hybrid}
\begin{algorithmic}[1]
\Require premises $P$, hypothesis $h$, LLM $M$, prover $\vdash$
\State Generate proof tree $T=\{(S_i,d_i)\}_{i=1}^{k}$ and raw label $r$
\State $\Dval \gets \emptyset$
\For{each derivation $(S_i,d_i)$ in $T$}
  \If{$S_i \vdash d_i$}
    \State $\Dval \gets \Dval \cup \{d_i\}$
  \EndIf
\EndFor
\If{$r \in \{\textsc{True},\textsc{False}\}$}
  \State \Return $r$ \Comment{preserve confident verdict}
\EndIf
\State $P' \gets P \cup \Dval$
\If{$P' \vdash h$}
  \State \Return \textsc{True}
\ElsIf{$P' \vdash \neg h$}
  \State \Return \textsc{False}
\Else
  \State \Return \textsc{Uncertain}
\EndIf
\end{algorithmic}
\end{algorithm}

\section{Experimental Setup}
\label{sec:setup}

\paragraph{Datasets.} We use balanced subsets of four logical-NLI benchmarks: \textbf{FOLIO} \citep{han-etal-2024-folio} (182 expert-written, three-label examples), \textbf{ProofWriter} \citep{tafjord2021proofwriter} (225, open-world, depth-controlled), \textbf{ProverQA} \citep{qi2025proverqa} (210, the hard split with 6--9 reasoning steps), and \textbf{PrOntoQA} \citep{PrOntoQA} (200, binary closed-world). The three open-world sets use the label space $\{\textsc{True},\textsc{False},\textsc{Uncertain}\}$.

\paragraph{Models.} We evaluate five instruction-tuned LLMs spanning vendors and scales: DeepSeek~V4~Flash~\citep{deepseekv42026}, GPT-5~Nano~\citep{singh2026openaigpt5card}, Gemini~3.1~Flash-Lite~\citep{gemini31flashlite}, Grok~4.3~\citep{grok4card}, and Qwen~3.5~Flash~\citep{yang2025qwen3technicalreport}. All methods use the same decoding configuration per model; the symbolic stages call Prover9 with a fixed timeout.

\paragraph{Methods.} We compare four \emph{internal} methods that we run under identical conditions: \textbf{Standard} (label only), \textbf{CoT} (short natural-language reasoning then label) \citep{wei2022chain}, \textbf{Autoformalization} (whole-problem first-order translation discharged by Prover9, the monolithic baseline), and \textbf{Hybrid Reasoning}. We also report three \emph{external} neurosymbolic baselines, \textbf{Logic-LM} \citep{pan-etal-2023-logic} and \textbf{LINC} \citep{olausson2023linc}, and the compositional \textbf{CLOVER} \citep{ryu2025divide}, on the splits where they apply (reproduction details in Appendix~\ref{app:external_repro}; execution rate, executable accuracy, and final accuracy reported separately in Appendix~\ref{app:executability}). For reference we also report the LLM-only \emph{Hybrid Raw} verdict (the Stage-1 prediction before validation and repair).
\begin{figure*}[!t]
  \centering
  \includegraphics[width=0.92\textwidth]{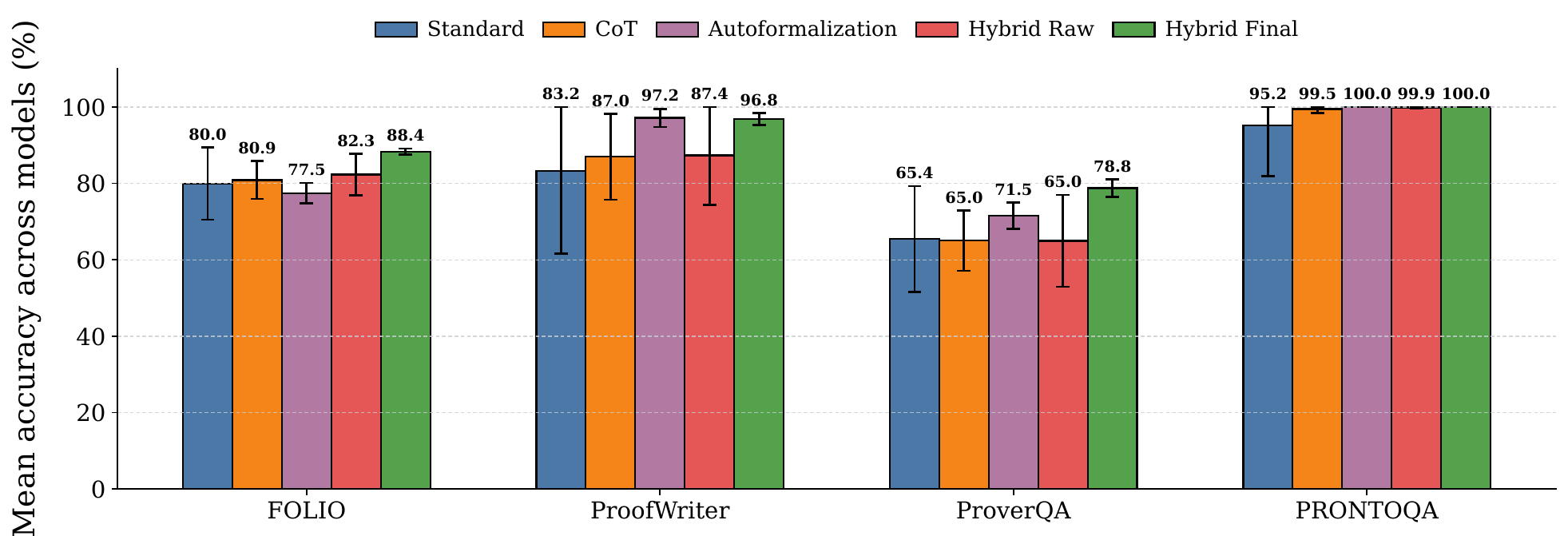}
  \caption{Mean accuracy across the five models for each internal method, with $95\%$ $t$-confidence intervals. \emph{Hybrid Final} is the validated, repaired pipeline; \emph{Hybrid Raw} is the LLM's Stage-1 verdict before validation and repair. The raw verdict is not reliably better than CoT, but validation and uncertainty-gated repair lift it to the top, especially on ProverQA.}
  \label{fig:method}
\end{figure*}
\paragraph{Metrics.} The primary metric is accuracy. For the repair stage we report how often it fires, its precision, and its net effect (helpful minus harmful relabellings; Table~\ref{tab:repair}). Per-class precision and recall are in Appendix~\ref{app:perlabel}.

\section{Results}
\label{sec:results}

\begin{table*}[t]
  \centering
  \small
  \setlength{\tabcolsep}{5pt}
  \begin{tabular}{ll ccc cc c c}
    \toprule
    & & \multicolumn{3}{c}{\emph{External}} & \multicolumn{4}{c}{\emph{Internal (this work's harness)}} \\
    \cmidrule(lr){3-5}\cmidrule(lr){6-9}
    Dataset & Model & Logic-LM & LINC & CLOVER & Standard & CoT & Autoform. & Hybrid \\
    \midrule
    \multirow{6}{*}{\textbf{FOLIO}}
      & DeepSeek V4 Flash      & 83.5 & 81.3 & 75.3 & \snd{84.6} & 82.4 & 79.1 & \best{88.5} \\
      & GPT-5 Nano             & 78.0 & 77.5 & 78.0 & \snd{79.1} & 76.4 & 74.7 & \best{87.4} \\
      & Gemini 3.1 Flash-Lite  & \snd{80.8} & 74.2 & 75.8 & 67.0 & 76.9 & 78.0 & \best{89.0} \\
      & Grok 4.3               & \snd{85.7} & 75.8 & 74.7 & 84.1 & 85.2 & 79.7 & \best{88.5} \\
      & Qwen 3.5 Flash         & \snd{87.4} & 81.3 & 76.4 & 85.2 & 83.5 & 75.8 & \best{88.5} \\
      \cmidrule(l){2-9}
      & \emph{Mean}            & \emph{\snd{83.1}} & \emph{78.0} & \emph{76.0} & \emph{80.0} & \emph{80.9} & \emph{77.5} & \emph{\textbf{88.4}} \\
    \midrule
    \multirow{6}{*}{\textbf{ProofWriter}}
      & DeepSeek V4 Flash      & 83.6 & \snd{96.0} & 95.1 & 89.8 & 87.1 & 95.6 & \best{96.4} \\
      & GPT-5 Nano             & 78.7 & 97.3 & 96.0 & 96.4 & 96.0 & \best{99.1} & \snd{98.7} \\
      & Gemini 3.1 Flash-Lite  & 78.7 & \best{97.3} & \snd{95.1} & 53.3 & 72.9 & 94.7 & \snd{95.1} \\
      & Grok 4.3               & 81.3 & 96.0 & 86.2 & 83.6 & 85.3 & \best{98.2} & \snd{96.9} \\
      & Qwen 3.5 Flash         & 85.8 & 95.6 & 95.6 & 92.9 & 93.8 & \best{98.2} & \snd{96.9} \\
      \cmidrule(l){2-9}
      & \emph{Mean}            & \emph{81.6} & \emph{96.4} & \emph{93.6} & \emph{83.2} & \emph{87.0} & \emph{\textbf{97.2}} & \emph{\snd{96.8}} \\
    \midrule
    \multirow{6}{*}{\textbf{ProverQA}}
      & DeepSeek V4 Flash      & \snd{74.8} & 70.0 & 71.0 & 71.0 & 71.4 & 73.8 & \best{80.0} \\
      & GPT-5 Nano             & 68.1 & 63.8 & \snd{71.9} & 59.0 & 59.0 & 69.5 & \best{76.2} \\
      & Gemini 3.1 Flash-Lite  & \snd{71.9} & 67.6 & 70.0 & 49.0 & 59.0 & 68.1 & \best{78.1} \\
      & Grok 4.3               & \snd{79.0} & 73.8 & 71.9 & 71.9 & 63.8 & 74.8 & \best{81.0} \\
      & Qwen 3.5 Flash         & 75.7 & 48.6 & \snd{78.1} & 76.2 & 71.9 & 71.4 & \best{78.6} \\
      \cmidrule(l){2-9}
      & \emph{Mean}            & \emph{\snd{73.9}} & \emph{64.8} & \emph{72.6} & \emph{65.4} & \emph{65.0} & \emph{71.5} & \emph{\textbf{78.8}} \\
    \midrule
    \multirow{6}{*}{\textbf{PrOntoQA}}
      & DeepSeek V4 Flash      & 98.0 & \best{100.0} & \best{100.0} & \best{100.0} & \snd{99.5} & \best{100.0} & \best{100.0} \\
      & GPT-5 Nano             & 93.5 & \best{100.0} & \snd{99.5} & \best{100.0} & \best{100.0} & \best{100.0} & \best{100.0} \\
      & Gemini 3.1 Flash-Lite  & \snd{99.0} & \best{100.0} & \best{100.0} & 76.0 & 98.0 & \best{100.0} & \best{100.0} \\
      & Grok 4.3               & \snd{98.5} & \best{100.0} & \best{100.0} & \best{100.0} & \best{100.0} & \best{100.0} & \best{100.0} \\
      & Qwen 3.5 Flash         & \best{100.0} & \best{100.0} & \best{100.0} & \best{100.0} & \best{100.0} & \best{100.0} & \best{100.0} \\
      \cmidrule(l){2-9}
      & \emph{Mean}            & \emph{97.8} & \emph{\textbf{100.0}} & \emph{\snd{99.9}} & \emph{95.2} & \emph{99.5} & \emph{\textbf{100.0}} & \emph{\textbf{100.0}} \\
    \bottomrule
  \end{tabular}
  \caption{Accuracy (\%) by dataset and model. \best{Bold} marks the best method in each row, \snd{underline} the second best. \emph{Autoform.}\ is whole-problem autoformalization discharged by Prover9 (the monolithic neurosymbolic baseline); \emph{Hybrid} is the full pipeline. PrOntoQA is saturated and serves as a sanity check.}
  \label{tab:main}
\end{table*}

\paragraph{FOLIO and ProverQA.} On the two hardest, non-saturated benchmarks, Hybrid Reasoning has the highest accuracy with every model  (Table~\ref{tab:main}). On FOLIO it improves the mean over the strongest competing method (Logic-LM, $83.1$) to $88.4$, peaking at $89.0$ (Gemini~3.1~Flash-Lite). On ProverQA it raises the mean from $73.9$ (Logic-LM) to $78.8$, peaking at $81.0$ (Grok~4.3). Averaged over models, Hybrid's lead over every baseline on both datasets is significant (\S\ref{sec:significance}). These are the regimes where monolithic translation often fails to execute and where CoT is least reliable: long chains, three-way labels, and semantically varied premises.

\paragraph{ProofWriter and PrOntoQA.} ProofWriter is comparatively easy for whole-problem formalisation, and there the monolithic \emph{Autoformalization} baseline is marginally strongest on average ($97.2$ vs.\ Hybrid's $96.8$); Hybrid is best or second on every model and never trails the best symbolic method by more than $2.2$ points. On PrOntoQA, Hybrid reaches 100\% with all five models, as do LINC and Autoformalization.

\paragraph{Where the gains come from.} The raw proof-tree verdict is barely better than CoT (Figure~\ref{fig:method}); validation and gated repair supply most of the gain, most visibly on ProverQA (raw $65.0$, final $78.8$; \S\ref{sec:repair-analysis} and Appendix~\ref{app:ablations}).

\section{Analysis}
\label{sec:analysis}
\subsection{Statistical significance}
\label{sec:significance}

Each benchmark has $182$--$225$ examples, so we test every gap on paired examples. Per model, we use a two-sided exact McNemar test \citep{mcnemar}. For the mean over models, each example contributes its Hybrid-minus-baseline correctness difference averaged over the five models; $p$ comes from a two-sided sign-flip permutation test with $100{,}000$ flips \citep{dror}, and the $95\%$ CI from $10{,}000$ paired bootstrap resamples \citep{bootstrap}. Holm correction \citep{holm} is applied within three families: \emph{primary} (FOLIO, ProverQA), \emph{repair} (Final vs.\ Raw), and \emph{secondary} (ProofWriter, PrOntoQA). Appendix~\ref{app:stats} gives details and all per-model results.

Averaged over the five models, Hybrid's lead over every baseline on FOLIO and ProverQA is significant (Table~\ref{tab:sig}; Holm-corrected $p<.001$). The smallest leads are over Logic-LM: $+5.3$ points on FOLIO ($95\%$ CI $[2.3, 8.5]$) and $+4.9$ on ProverQA ($[2.4, 7.3]$). The per-model evidence is weaker. Hybrid is ahead of Logic-LM on all ten model-dataset pairs, but only GPT-5~Nano on FOLIO remains significant after Holm correction (Appendix~\ref{app:stats}), so our significance claims concern the mean over models.

On ProofWriter, neither average gap to the strongest symbolic baselines is significant: $-0.4$ points against Autoformalization ($95\%$ CI $[-1.2, 0.4]$) and $+0.4$ against LINC ($[-0.6, 1.4]$), both with Holm-corrected $p=1.0$, while Hybrid stays significantly ahead of Logic-LM, CLOVER, Standard, and CoT ($p<.001$). Hybrid Final's gain over Hybrid Raw is significant on FOLIO, ProofWriter, and ProverQA (Holm-corrected $p<.001$; ProverQA $+13.8$ points, $95\%$ CI $[10.4, 17.3]$), but not on saturated PrOntoQA ($+0.1$, $p=1.0$).

\begin{table}[t]
  \centering
  \small
  \setlength{\tabcolsep}{4pt}
  \begin{tabular}{l rc rc}
    \toprule
    & \multicolumn{2}{c}{FOLIO} & \multicolumn{2}{c}{ProverQA} \\
    \cmidrule(lr){2-3}\cmidrule(lr){4-5}
    Hybrid vs. & $\Delta$ & 95\% CI & $\Delta$ & 95\% CI \\
    \midrule
    Logic-LM   & $+5.3$  & $[2.3, 8.5]$  & $+4.9$  & $[2.4, 7.3]$ \\
    LINC       & $+10.3$ & $[5.8, 15.1]$ & $+14.0$ & $[10.2, 17.7]$ \\
    CLOVER     & $+12.3$ & $[8.2, 16.6]$ & $+6.2$  & $[3.2, 9.1]$ \\
    Standard   & $+8.4$  & $[5.9, 11.0]$ & $+13.3$ & $[9.8, 16.8]$ \\
    CoT        & $+7.5$  & $[5.2, 9.9]$  & $+13.7$ & $[10.0, 17.5]$ \\
    Autoform.  & $+10.9$ & $[6.7, 15.2]$ & $+7.2$  & $[4.1, 10.5]$ \\
    \midrule
    Hybrid Raw & $+6.0$  & $[4.0, 8.5]$  & $+13.8$ & $[10.4, 17.3]$ \\
    \bottomrule
  \end{tabular}
  \caption{Paired tests of Hybrid against each method on the two hardest datasets, averaged over the five models. $\Delta$: Hybrid's accuracy minus the method's (points); \emph{95\% CI}: paired bootstrap interval over examples. All Holm-corrected sign-flip $p$-values are below $.001$. \emph{Hybrid Raw} is the Stage-1 verdict and forms its own correction family. Tests on ProofWriter and PrOntoQA and per-model McNemar tests are in Appendix~\ref{app:stats}.}
  \label{tab:sig}
\end{table}
\subsection{Repair precision and coverage}
\label{sec:repair-analysis}
Table~\ref{tab:repair} reports direct repair coverage and precision. Repair fires on a minority of instances (from $6.9\%$ on FOLIO to $19.4\%$ on ProverQA) and is correct 78--98\% of the time. Because repair is gated to the \textsc{Uncertain} region, it can harm only instances whose gold label is Uncertain. Across all datasets it makes 326 helpful and 19 harmful relabellings (net $+307$). The net is positive on every dataset, and the accuracy gain reaches +13.8 points on ProverQA. Repair fires most on ProverQA, where the model abstains on $47.4\%$ of decidable instances. FOLIO and ProofWriter have similar abstention rates ($19.7\%$ and $18.7\%$), but repair relabels about half of FOLIO's abstentions against $77\%$ of ProofWriter's (Appendix~\ref{app:err-abstain}).

\begin{table}[t]
  \centering
  \small
  \setlength{\tabcolsep}{4pt}
  \begin{tabular}{l ccc c c}
    \toprule
    Dataset & Fires & Rate & Prec. & Net & Gain \\
    \midrule
    FOLIO       & 63  & 6.9\%  & 90.5\% & $+55$  & $+6.0$ \\
    ProofWriter & 110 & 9.8\%  & 98.2\% & $+106$ & $+9.4$ \\
    ProverQA    & 204 & 19.4\% & 78.4\% & $+145$ & $+13.8$ \\
    PrOntoQA    & 1   & 0.1\%  & 100.0\% & $+1$  & $+0.1$ \\
    \bottomrule
  \end{tabular}
  \caption{Direct uncertainty-gated repair coverage and precision (pooled over models). \emph{Fires}: repair invocations; \emph{Rate}: firing rate; \emph{Prec.}: fraction of firings that reach the gold label; \emph{Net}: helpful minus harmful relabellings among repair firings; \emph{Gain}: raw$\to$final accuracy change (points).}
  \label{tab:repair}
\end{table}

\begin{figure*}[t]
  \centering
  \includegraphics[width=0.92\textwidth]{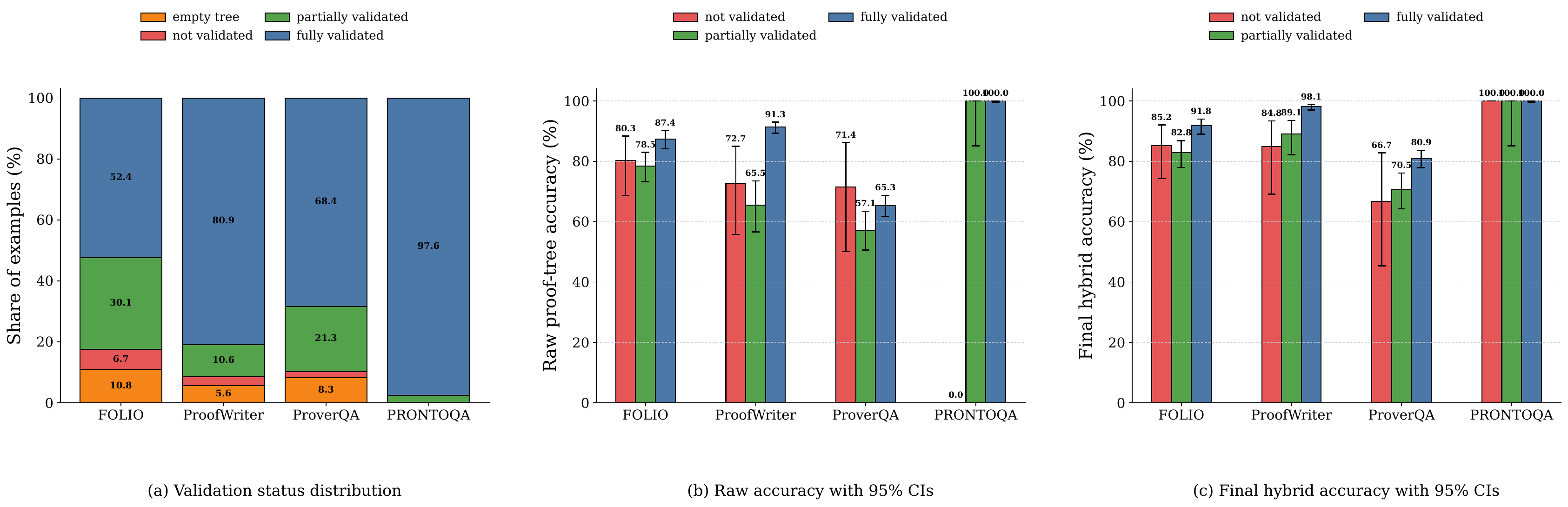}
  \caption{Validation-status analysis, pooled across models. (a) Share of instances by how much of the proof tree validated. (b) The LLM's \emph{raw} accuracy is higher for fully than for partially validated trees, making validation status a label-free reliability signal. (c) The repaired pipeline improves accuracy at every level, most where raw reliability is lowest. Error bars are $95\%$ Wilson intervals.}
  \label{fig:validation}
\end{figure*}

\subsection{Ablations}
Appendix~\ref{app:ablations} varies the two design choices of Stage~3 while holding the Stage-1 outputs fixed: which derivations enter the repair premise set (none, all proposed derivations, or only $\Dval$), and whether repair may revise every raw verdict or only \textsc{Uncertain} ones.
Gated repair from $P\cup\Dval$, the configuration used throughout, has the highest mean accuracy of the six combinations ($91.0$).
Repairing without derivations lowers accuracy on FOLIO and ProverQA and slightly raises it on ProofWriter.
Admitting unvalidated derivations lowers mean accuracy by more than two points, largely on Gemini~3.1~Flash-Lite: the prover then resolves correct raw \textsc{Uncertain} verdicts to \textsc{True} or \textsc{False} ($89$ such regressions under the gate, against $19$ with $\Dval$).
Without the gate, repair overturns only $10$--$15$ of the $2{,}500$ decisive raw verdicts, for a net loss of $2$--$5$ correct predictions.

\subsection{Validation status and raw accuracy}\label{sec:validation-depth}
Figure~\ref{fig:validation} stratifies instances by validation status, pooled across models (per-model gains in Appendix~\ref{app:permodel}). The distribution (panel a) shows that most proof trees validate fully or partially. In panel (b), raw accuracy is higher for fully than for partially validated trees: on FOLIO from $78.5\%$ (partially) to $87.4\%$ (fully), on ProofWriter from $65.5\%$ to $91.3\%$, and on ProverQA from $57.1\%$ to $65.3\%$. In other words, how much of a proof tree a prover can independently discharge is a usable confidence signal for the LLM's verdict, available at inference with no labels. Panel (c) shows the repaired pipeline lifts accuracy at every validation level, with the largest gains at the levels where raw accuracy is lowest. Monolithic pipelines, which emit a single global program, expose no comparable signal.

\subsection{Where repair moves predictions}
\label{sec:label-level}
\begin{figure}[h]
  \centering
  \includegraphics[width=0.4\textwidth]{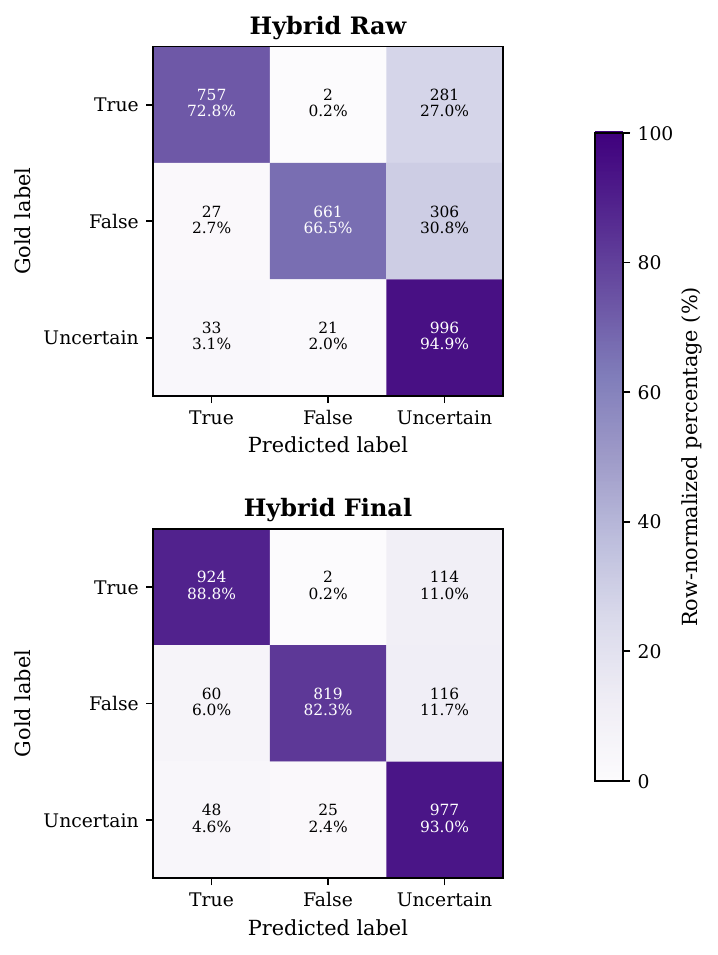}
  \caption{Row-normalised confusion before (Hybrid Raw) and after (Hybrid Final) repair, pooled over the three open-world datasets. Repair reclaims \textsc{True}/\textsc{False} cases the model had left \textsc{Uncertain} (recall $73{\to}89\%$ and $66{\to}82\%$).}
  \label{fig:confusion}
\end{figure}

Before repair, 27.0\% of gold-True and 30.8\% of gold-False instances are predicted Uncertain, against 0.2\% and 2.7\% given the opposite label (Figure~\ref{fig:confusion}). Repair resolves many of these abstentions, raising \textsc{True} recall from $72.8\%$ to $88.8\%$ and \textsc{False} recall from $66.5\%$ to $82.3\%$. Uncertain recall falls from 94.9\% to 93.0\%, and the share of gold-\textsc{False} instances predicted \textsc{True} rises from $2.7\%$ to $6.0\%$, mostly on ProverQA, where \textsc{True} precision drops from $88.9\%$ to $81.4\%$ (Appendix~\ref{app:perlabel}; label distributions in Appendix~\ref{app:prediction}). Against monolithic Autoformalization (Appendix~\ref{app:complementarity}), Hybrid recovers $77.7\%$ of the instances only Autoformalization gets right, preserves $95.0\%$ of those only the raw verdict gets right, and rescues $19.1\%$ of those both get wrong.

\subsection{Derivation economy}
Pooled over the open-world datasets, models propose $4.4$ derivations per instance, of which $4.1$ ($93\%$) validate; the retained fraction ranges from $85\%$ on FOLIO (see Cases C and D in Appendix~\ref{app:cases}) to $95\%$ on ProofWriter and ProverQA. Validation removes roughly 7\% of derivations on average, enough to change repair outcomes (Appendix~\ref{app:ablations}) without discarding most of what the model produced.

\subsection{Error analysis}
\label{sec:errors}
On decidable instances, the proof-tree stage abstains on $18.7$--$47.4\%$ (most on ProverQA) and gives the opposite label on at most $3.1\%$. Abstentions rise with proof depth, and on ProverQA abstaining chains are mostly valid but about half as long as successful ones. The prover rejects far more steps that compress several inferences or assert a paraphrase ($21.5$--$28.7\%$) than single-rule steps ($2.8\%$ for modus ponens). Repair resolves $48$--$77\%$ of abstentions; $41$ of its $52$ wrong relabellings come from contradictory formalizations, which let the prover prove anything. Appendix~\ref{app:cases} analyses each pattern with worked cases.

\section{Discussion}
\label{sec:discussion}

\paragraph{Abstentions.} When the raw verdict is wrong on a \textsc{True} or \textsc{False} instance, it is almost always \textsc{Uncertain} (\S\ref{sec:label-level}), even though $93\%$ of the proposed derivations pass validation. The model tends to propose sound steps and stop early (Appendix~\ref{app:err-abstain}). Since it seldom commits to the wrong label, the gate blocks few useful corrections. Ungated repair overturns only $13$ of $2{,}500$ decisive verdicts and breaks more correct ones than it fixes (Appendix~\ref{app:ablations}).

\paragraph{Validated derivations.} Repair from the premises alone already recovers most of the gain over the raw verdict, so part of the improvement comes from a second, symbolic attempt on abstentions. The validated derivations add to it on FOLIO and ProverQA but not on ProofWriter (Table~\ref{tab:ablation_permodel}), which monolithic Autoformalization nearly solves. We hypothesize that they help when a problem is hard to formalize in one pass, by supplying intermediate facts that a single global translation misses.

\paragraph{Local and global guarantees.}
Hybrid certifies validated derivations and repaired labels but not decisive raw verdicts. For those, Hybrid still provides a partial validation signal: the share of the proof tree that validates indicates how far the verdict can be trusted (\S\ref{sec:validation-depth}). Whole-problem formalization certifies each label it proves, though only relative to its translation, and its executed predictions can still be wrong (Appendix~\ref{app:executability}). It suits users who need a certificate; Hybrid suits those who need accuracy (Table~\ref{tab:sig}).

\paragraph{Sensitivity to the backbone.} Admitting unvalidated derivations hurts mainly Gemini~3.1~Flash-Lite, the model whose derivations fail validation most often. It loses $7.1$--$13.8$ points on the open-world datasets, while no other model loses more than $4.0$ (Table~\ref{tab:ablation_permodel}). Pooled over models, $72$ of the $89$ harmful relabellings in this variant involve a rejected derivation (Appendix~\ref{app:ablations}). With validation, final accuracy stays within $4.8$ points across models on each open-world dataset, against up to $16.5$ with all derivations and $23.1$ for CoT (Tables~\ref{tab:main} and~\ref{tab:ablation_permodel}).

\paragraph{Remaining errors.} Repair is least precise on ProverQA, where most wrong relabellings come from contradictory formalized premises (e.g., an exclusive ``either \ldots\ or'' read as inclusive), from which the prover can prove any goal (\S\ref{sec:errors}). Since the prover's guarantee covers only the formulas it receives, further gains will most likely come from more faithful formalization.

\section{Conclusion}
\label{sec:conclusion}
We presented Hybrid Reasoning, in which a theorem prover checks each step of an LLM's proof tree and takes over the instances the model leaves \textsc{Uncertain}. Averaged over five LLMs, it is significantly more accurate than every baseline on FOLIO and ProverQA and matches the best neurosymbolic baselines on ProofWriter and PrOntoQA. Repair fixes many more errors than it introduces without changing decisive verdicts. The share of a proof tree that validates also indicates, without labels, how far the model's own verdict can be trusted. By checking the model's reasoning step by step and deferring to it whenever it commits to a label, a prover can add to what an LLM gets right and stay useful on problems that cannot be formalized correctly as a whole.

\section*{Limitations}

\textit{Scope of logic.} Our symbolic stage targets first-order entailment, and we evaluate it only on first-order benchmarks. Temporal, modal, defeasible, and arithmetic reasoning would each need its own translator and solver; the three-stage structure could carry over to them, but we have not tested this. Our results therefore do not establish that the method helps where reasoning is less local or harder to formalize.

\textit{Autoformalization bottleneck.} Validation can fail silently
when a derivation is translated into syntactically valid but
semantically incorrect first-order logic. Similar
failure modes have been observed in prior translation-based systems
\citep{olausson2023linc, pan-etal-2023-logic}, and they are intrinsic to
any translation-based gate; they bound the precision of our repair
stage.

\textit{Conservative repair policy.} The gate fires only when the
raw verdict is \textsc{Uncertain} and refuses to overwrite confident
\textsc{True}/\textsc{False} predictions. This asymmetry underwrites
our precision numbers but caps recoverable coverage on confidently
wrong instances. More aggressive policies, such as
confidence-thresholded overwrites, staged arbitration, or learned
risk controls, would shift the precision--coverage frontier and
warrant separate evaluation.

\textit{Validation depth as a discarded training signal.} Raw accuracy is higher for fully than for partially validated trees (\S\ref{sec:validation-depth}),
but we use this signal only at inference time. The validated set
$\mathcal{D}^+$ is naturally-labeled step-level data and could
supervise process reward models, or be distilled into a single
forward pass that approximates the gating behaviour at lower cost.

\textit{Inference cost.} Cost scales with proof-tree size, since
each derivation requires an autoformalization and prover call in
Stage~2, and rises further for instances whose raw verdict is
\textsc{Uncertain}, which trigger an additional repair pass in
Stage~3. Confidently-labeled instances with short trees are
relatively cheap; long-chain \textsc{Uncertain} instances incur the
full pipeline. Sharing one formalization across a problem’s derivations, or caching prover calls, would reduce this cost.

\textit{Language.} All experiments are in English; multilingual
transfer would depend on both LLM translation quality and the
availability of benchmarks with comparable logical structure.

\section*{Ethics Statement}
This work studies logical reasoning over public benchmarks and involves no human subjects or personal data. The datasets are released for research, and our use is consistent with their intended purpose. Because the models are commercial APIs that may change or be retired, our results may not be exactly reproducible; we release our prompts and will release our code.

\bibliography{custom}

\appendix

\section{Dataset details}
\label{app:data}
 
We evaluate on balanced subsets of four logical-NLI benchmarks.
Every method, internal and external, is run on the identical
split for a given dataset, so all reported numbers are directly
comparable; all sampling uses a fixed random seed.
Each instance is stored as a natural-language premise set $P$ and a
hypothesis $h$ with a gold label, alongside reference first-order-logic
annotations where the source provides them.
Table~\ref{tab:datastats} summarises the splits.
 
\paragraph{FOLIO.} FOLIO is an expert-written
first-order-logic reasoning benchmark whose problems use naturalistic
language and a three-way label space
$\{\textsc{True},\textsc{False},\textsc{Uncertain}\}$.
We use the validation split after removing the $22$ examples that
\citet{olausson2023linc} identify as erroneous (unbalanced parentheses,
gold-label/FOL mismatches, or premise/FOL count mismatches), yielding
the same $182$-example set used by LINC.
We keep the dataset's original label space; the resulting label
distribution is mildly imbalanced ($63$ \textsc{True},
$54$ \textsc{False}, $65$ \textsc{Uncertain}).
FOLIO has the fewest premises per instance ($5.3$ on average).
 
\paragraph{ProofWriter.} ProofWriter is
a synthetically generated deductive benchmark with controllable proof
depth.
We use the open-world-assumption (OWA) portion, which preserves the
three-way label space, and draw a depth-stratified, label-balanced
subset of $225$ examples from the test split: $45$ examples at each
gold proof depth $d\in\{1,2,3,4,5\}$ and $75$ examples per label.
This stratification procedure follows the design used in recent
decompose-and-verify work \citep{quan2026decompose}; because we resample
under our own seed, the specific instances differ.
ProofWriter instances have many short premises ($15.3$ on average),
which makes whole-problem formalisation comparatively easy.
 
\paragraph{PrOntoQA.} PrOntoQA is a
synthetic deductive benchmark with a binary label space.
Following the configuration adopted by
Logic-LM \citep{pan-etal-2023-logic}---the hardest \emph{fictional-character},
five-hop subset---we take their released split and draw a
label-balanced subset of $200$ examples
($100$ \textsc{True}, $100$ \textsc{False}).
Because strong prompting nearly solves this setting
(\S\ref{sec:results}), we use PrOntoQA only to check that the added stages do not lower accuracy.
 
\paragraph{ProverQA.} ProverQA pairs
LLM-generated natural language with symbolic-prover-validated labels.
We use its \emph{hard} split, characterised by long chains
($6$--$9$ reasoning steps) and many premises ($18.4$ on average),
and draw a label-balanced subset of $210$ examples from the hard
validation set ($70$ per label).
This is the most demanding of the four benchmarks and is where Hybrid
Reasoning's gains are largest.
 
\begin{table}[h]
  \centering\small
  \setlength{\tabcolsep}{4pt}
  \begin{tabular}{l c c c c}
    \toprule
    Dataset & $N$ & Labels & \#Prem. & Depth \\
    \midrule
    FOLIO       & 182 & 3 & 5.3  & ---  \\
    ProofWriter & 225 & 3 & 15.3 & 1--5 \\
    PrOntoQA    & 200 & 2 & 17.3 & 5    \\
    ProverQA    & 210 & 3 & 18.4 & 6--9 \\
    \bottomrule
  \end{tabular}
  \caption{Evaluation splits. $N$: examples; \emph{Labels}: size of the
    label space (3\,=\,\textsc{True}/\textsc{False}/\textsc{Uncertain},
    2\,=\,binary); \emph{\#Prem.}: mean premises per instance;
    \emph{Depth}: gold reasoning depth (where defined).
    ProofWriter is depth-balanced ($45$ per depth) and all
    three-label sets are label-balanced except FOLIO, which keeps
    its natural distribution.}
  \label{tab:datastats}
\end{table}
 
\section{Model and decoding details}
\label{app:models}
 
We used five hosted instruction-tuned backbone models: DeepSeek~V4~Flash,
Gemini~3.1~Flash-Lite, GPT-5~Nano, Grok~4.3, and Qwen~3.5~Flash.
For each backbone, we fixed a single reasoning configuration and used
it consistently across all methods that call that model:
Standard prompting, CoT prompting, Autoformalization, Hybrid Raw,
Hybrid Final, and the reproduced external baselines whenever the
baseline implementation exposed the same controls.
The purpose of this design is to isolate the effect of the reasoning
procedure rather than confounding it with model-specific decoding
choices or per-method tuning.
 
Table~\ref{tab:model_settings} reports the backbone configurations.
DeepSeek~V4~Flash was run with reasoning enabled and reasoning effort
set to high.
Gemini~3.1~Flash-Lite was run with the provider default configuration
used by the API, which leaves the model's default thinking behaviour
active.
GPT-5~Nano was run with reasoning effort set to high.
Grok~4.3 was run with reasoning effort set to low.
Qwen~3.5~Flash was run with thinking enabled.
These settings were chosen before evaluation began and were not changed
by dataset, prompt type, or method.
 
\begin{table}[h]
  \centering\small
  \setlength{\tabcolsep}{4pt}
  \begin{tabular}{ll}
    \toprule
    Backbone model & Fixed reasoning configuration \\
    \midrule
    DeepSeek V4 Flash     & reasoning enabled; effort high         \\
    Gemini 3.1 Flash-Lite & provider default thinking behaviour     \\
    GPT-5 Nano            & reasoning effort high                   \\
    Grok 4.3              & reasoning effort low                    \\
    Qwen 3.5 Flash        & thinking enabled                        \\
    \bottomrule
  \end{tabular}
  \caption{Backbone model configurations used throughout the
    evaluation. Except for the LINC reproduction described below,
    temperature and top-$p$ were left unspecified and provider
    defaults were used.}
  \label{tab:model_settings}
\end{table}
 
For all internally implemented methods, we did not pass explicit
temperature or top-$p$ values.
This applies to the label-only baseline, CoT baseline, whole-problem
Autoformalization baseline, proof-tree generation stage, and
repair-related model calls.
We therefore treat sampling as part of the provider default
configuration, while keeping the higher-level reasoning or thinking
controls fixed as shown in Table~\ref{tab:model_settings}.
 
The only exception is the LINC reproduction.
The original LINC codebase relies on stochastic model sampling to
produce multiple formalisations before prover aggregation.
To preserve this intended behaviour, we retained LINC's stochastic
decoding design and used temperature~$0.8$ for all LINC runs whose
backbone API accepted an external temperature parameter.
GPT-5~Nano did not expose an external temperature setting in our
configuration; for this backbone, LINC therefore relied on the model's
provider-side stochasticity.
No additional decoding parameter, reasoning setting, or model-specific
prompt variant was tuned for Hybrid Reasoning.

\section{Executable symbolic reasoning}
\label{app:executability}

\begin{figure*}[t]
  \centering
  \includegraphics[width=0.92\textwidth]{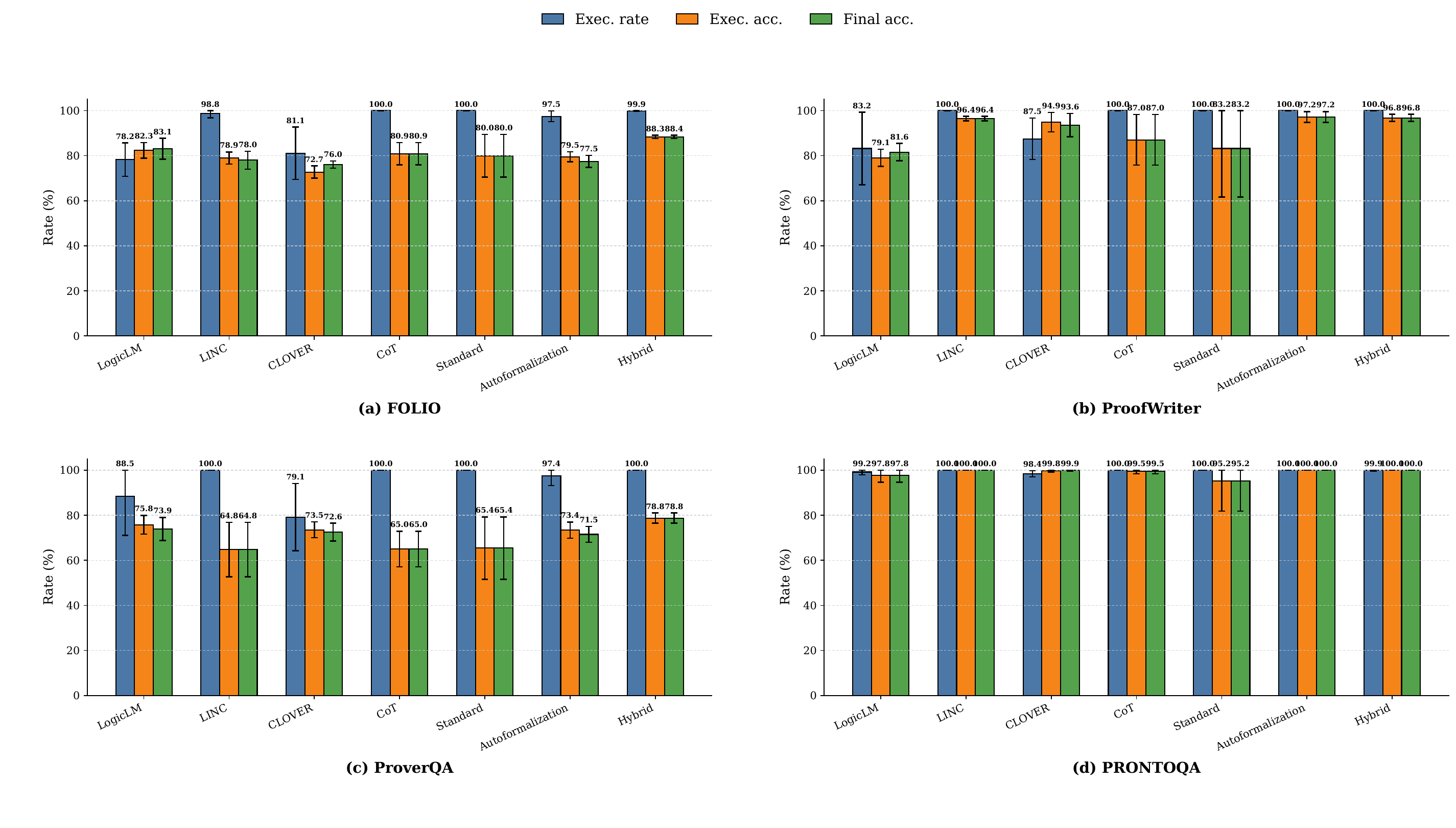}
  \caption{Execution rate, executable accuracy, and final accuracy by
  dataset and method. Execution rate measures the fraction of instances
  for which the method obtains an executable symbolic result under its
  own implementation. Executable accuracy is accuracy restricted to those
  executable cases; final accuracy includes the method's complete
  decision procedure, including fallback behavior where applicable.}
  \label{fig:exec_rates}
\end{figure*}

Figure~\ref{fig:exec_rates} separates execution rate, executable
accuracy, and final accuracy. Execution failures generally correspond to
cases where a method cannot obtain a usable structured result, such as a
parse failure or a critical formalization error that prevents a prover
decision. This distinction is useful for interpreting full-translation
baselines: Logic-LM and CLOVER have lower execution rates on FOLIO and ProofWriter,
while LINC executes more consistently, but has the lowest ProverQA accuracy of the three (64.8), so executable output is not necessarily faithful. Hybrid’s execution depends mainly on obtaining a parseable proof tree output, after which local
derivation validation and gated repair determine the final decision.

\section{External baseline reproduction details}
\label{app:external_repro}

We reproduced Logic-LM~\citep{pan-etal-2023-logic}, LINC~\citep{olausson2023linc} and CLOVER \citep{ryu2025divide} using their publicly available codebases.
Inputs were serialized from our premise--hypothesis records, and outputs
were normalized into the corresponding benchmark label space:
\textsc{True}/\textsc{False}/\textsc{Uncertain} for the open-world
datasets and binary \textsc{True}/\textsc{False} for PrOntoQA.

\paragraph{Logic-LM.}
Logic-LM was reproduced following its original solver paths: Prover9 for
FOLIO, and Pyke \citep{frederiksen2008pyke} for PrOntoQA and ProofWriter. Since Logic-LM did not
include a ProverQA evaluation path, we converted ProverQA examples into
the same input format used by its FOLIO pipeline and ran them with the
Prover9 solver. We used the strongest setting from the Logic-LM
implementation: CoT fallback enabled and self-refinement disabled.

We made a minimal prompt-compatibility patch for modern instruction-tuned
models. The original Logic-LM prompts often led recent models to answer
the problem directly or add free-form explanatory text, producing outputs
that were not parseable by the existing codebase. We therefore kept the
few-shot examples, symbolic solver choices, and downstream parser
interfaces unchanged, but added explicit final-output constraints to the
last query. The following boxes list the exact additions. The patch was
intended only to preserve the original completion format expected by
Logic-LM's parser.

\paragraph{Logic-LM prompt-compatibility patch.}\mbox{}\par\vspace{0.35em}

\begin{promptbox}{Logic-LM CoT answer patch: FOLIO}
For the final example, output only:
1) a short reasoning continuation under "Reasoning:"
2) one final line in exactly this format: The correct option is: <A/B/C>
Do not add headings, bullet points, markdown, or extra answer lines.
\end{promptbox}

\begin{promptbox}{Logic-LM CoT answer patch: PrOntoQA}
For the final example, output only:
1) a short reasoning continuation under "Reasoning:"
2) one final line in exactly this format: The correct option is: <A/B>
Do not add headings, bullet points, markdown, or extra answer lines.
\end{promptbox}

\begin{promptbox}{Logic-LM CoT answer patch: ProofWriter}
For the final example, output only:
1) a short reasoning continuation under "Reasoning:"
2) one final line in exactly this format: The correct option is: <A/B/C>
Do not add headings, bullet points, markdown, or extra answer lines.
\end{promptbox}

\begin{promptbox}{Logic-LM FOL formalization patch: FOLIO}
For the final problem, output only the formalization for that problem.
Do not explain your reasoning.
Do not answer whether the statement is true, false, or uncertain.
Do not repeat the previous examples.
Use exactly these three section headers in this order and nothing else:
Predicates:
Premises:
Conclusion:
\end{promptbox}

\begin{promptbox}{Logic-LM rule/fact formalization patch: PrOntoQA}
For the final problem, output only the formalization for that problem.
Do not explain your reasoning.
Do not answer whether the statement is true or false.
Do not repeat the previous examples.
Use exactly these four section headers in this order and nothing else:
Predicates:
Facts:
Rules:
Query:
\end{promptbox}

\begin{promptbox}{Logic-LM rule/fact formalization patch: ProofWriter}
For the final problem, output only the formalization for that problem.
Do not explain your reasoning.
Do not answer whether the statement is true, false, or uncertain.
Do not repeat the previous examples.
Use exactly these four section headers in this order and nothing else:
Predicates:
Facts:
Rules:
Query:
\end{promptbox}

\paragraph{LINC.}
LINC was reproduced without prompt changes or codebase-level changes to
its reasoning procedure. We plugged in the evaluated backbone models and
kept its original sampling-and-aggregation protocol. The only model-side
change was the temperature handling described in Appendix~\ref{app:models}:
we used temperature~$0.8$ where the model API accepted an external
temperature argument, while GPT-5~Nano relied on provider-side
stochasticity.

LINC did not include ProverQA or PrOntoQA evaluation paths in the original
setup. We therefore converted those examples into the FOLIO-style input
format required by its existing pipeline and ran them through the same
Prover9-based path.

\paragraph{CLOVER.}
CLOVER was reproduced using its compositional parsing, component
accumulation, translation, and Z3-based \citep{z3} verification pipeline. We used
the counter-interpretation verification setting (\texttt{logic\_cp})
with CoT fallback enabled. The released codebase exposes AR-LSAT and
ZebraLogic evaluation paths rather than paths for our four benchmarks.
We therefore converted our examples into the input format required by
the existing pipeline and implemented the theory-estimation stage
described in the paper, which generates a shared signature of entities
and predicates from the premises and hypothesis.

We retained the original parsing, accumulation, and verification
demonstrations, and supplied FOLIO-based translation demonstrations for
the adapted dataset paths. As with Logic-LM, we added explicit
final-output constraints where instruction-tuned models produced text
that the existing parsers could not use. These additions specify the
expected section headers, sentence--formula correspondence, declaration
names, and supported formula syntax.

\paragraph{CLOVER theory estimation and prompt-compatibility patches.}
\mbox{}\par\vspace{0.35em}

\begin{promptbox}{CLOVER theory-estimation prompt}
### Instruction
Given one problem in natural language, induce the first-order logic signature shared by all premises and the conclusion.
Output declarations only. Do not translate the premises or conclusion.
Use one entity sort named entities.
Declare every named individual as a constant in entities.
Declare every predicate needed by the premises and conclusion as a Function ending in bool.
Use stable, concise PascalCase predicate names and lowercase/camelcase entity names.
The example declarations are demonstration-only. Do not reuse example predicate names unless the target problem contains the same concept.
------
# Premises
Lawton Park is a neighbourhood in Seattle.
All citizens of Lawton Park use the zip code 98199.
Tom is a citizen of Lawton Park.
Daniel uses the zip code 98199.
# Conclusion
Tom uses the zip code 98199.
# Declarations
entities = EnumSort([daniel, lawtonPark, number98199, seattle, tom, anon_1, anon_2, anon_3])
Citizenof = Function([entities, entities] -> [bool])
Neighbourhoodin = Function([entities, entities] -> [bool])
Usezipcode = Function([entities, entities] -> [bool])
------
END OF DEMONSTRATIONS

Important output rules:
- Output only declaration lines.
- The first line must be: entities = EnumSort([...])
- Every other line must be: PredicateName = Function([entities, ...] -> [bool])
- Include predicates for the conclusion even if they appear nowhere in the premises.
- Keep one shared predicate name for repeated meanings.
- Do not copy demonstration predicates.
- Do not output formulas, explanations, markdown, bullets, or examples.
------
# Premises
[[CONTEXT]]
# Conclusion
[[QUERY]]
# Declarations
\end{promptbox}

The compatibility additions below follow the original demonstrations
and precede the final input, introduced by \texttt{END OF DEMONSTRATIONS}.

\begin{promptbox}{CLOVER dependency-parsing patch}
Important output rules:
- Use only the final input below.
- Do not copy demonstration sentences, declarations, or structures.
- Generate one or more possible logical dependency structures for the final sentence.
- Every possible structure must start with a numbered header such as ### 1:, ### 2:, etc.; use ### 1: even when there is only one structure.
- Keep the same component notation as the demonstrations: U for logic units, C for logic couplers, and D for logic dependents.
- Keep the same dependency notation: X -> Y, separated by semi-colons when there are multiple dependencies.
\end{promptbox}

\begin{promptbox}{CLOVER component-accumulation patch}
Important output rules:
- Use only the final sentence/structure pairs below.
- Do not copy demonstration sentences, structures, or accumulations.
- For each final # Sentence and # Structure pair, output the corresponding # Accumulation.
- If the structure is None, copy the sentence as the only accumulated line.
- Preserve quoted names and punctuation from the final sentence when they appear in accumulated lines.
- Keep the same output format as the demonstrations.
\end{promptbox}

\begin{promptbox}{CLOVER translation patch}
Important output rules:
- Only translate the sentence blocks shown immediately above.
- Do not repeat examples from this prompt.
- For every accumulated natural-language line, copy that line verbatim in the translation line.
- Every translation line must have exactly this format:
<verbatim accumulated line> ::: <FOL formula>
- Use only the declared sort "entities" for quantified variables, for example ForAll([x:entities], ...).
- Use only declared predicates/functions and keep predicate names exactly as declared.
- Do not invent new predicate names or rename declared predicates.
- Use declared entity names exactly; do not change spelling, casing, or underscores.
- Do not use Iff(...); translate if-and-only-if as And(Implies(A, B), Implies(B, A)).
- Use Exists([x:entities], And(...)) when an existential statement has multiple conditions; use Exists([x:entities], Predicate(x)) for a single condition.
\end{promptbox}

\begin{promptbox}{CLOVER verification patch}
Important output rules:
- Use only the final input below.
- Provide an Analysis section followed by an Answer section.
- The Answer section must be exactly Yes or No.
- Judge logical consistency using the declarations, solution, and sentence.
\end{promptbox}

\paragraph{Execution and final predictions.}
For all external baselines, execution follows the semantics of the
original implementation: an instance is executable when the generated
structured formalization can be parsed and passed through the relevant
solver path to obtain a usable symbolic result. Final accuracy includes
the method's complete decision procedure, including fallback behavior
where applicable. This is why Appendix~\ref{app:executability} reports
execution rate, executable accuracy, and final accuracy separately.

\section{Per-label precision and recall}
\label{app:perlabel}
 
Table~\ref{tab:perlabel} reports per-label precision and recall for
the monolithic \emph{Autoformalization} baseline, \emph{Hybrid Raw},
and \emph{Hybrid Final}.
On the open-world datasets, the main effect of repair is higher
\textsc{True} and \textsc{False} recall relative to Hybrid Raw,
confirming that the final system primarily recovers determinate cases
that the proof-tree stage leaves \textsc{Uncertain}.
PrOntoQA is binary, so the \textsc{Uncertain} columns are omitted.
 
\begin{table}[h]
  \centering\small
  \setlength{\tabcolsep}{3pt}
  \resizebox{\columnwidth}{!}{%
  \begin{tabular}{l l cc cc cc}
    \toprule
    Dataset & Method
      & \multicolumn{2}{c}{\textsc{True}}
      & \multicolumn{2}{c}{\textsc{False}}
      & \multicolumn{2}{c}{\textsc{Uncertain}} \\
    \cmidrule(lr){3-4}\cmidrule(lr){5-6}\cmidrule(lr){7-8}
    & & P & R & P & R & P & R \\
    \midrule
    \multirow{3}{*}{FOLIO}
      & Autoform.     & 90.7 & 71.4 & 98.8 & 61.9 & 66.6 & 96.3 \\
      & Hybrid Raw    & 88.3 & 81.3 & 94.7 & 72.2 & 72.2 & 91.7 \\
      & Hybrid Final  & 87.9 & 89.5 & 95.0 & 83.7 & 84.3 & 91.1 \\
    \midrule
    \multirow{3}{*}{ProofWriter}
      & Autoform.     & 100.0 & 95.7  & 100.0 & 95.7  & 92.1 & 100.0 \\
      & Hybrid Raw    & 99.4  & 82.1  & 100.0 & 80.3  & 72.8 & 99.7  \\
      & Hybrid Final  & 98.9  & 95.5  & 100.0 & 95.7  & 92.1 & 99.2  \\
    \midrule
    \multirow{3}{*}{ProverQA}
      & Autoform.     & 82.0 & 66.6 & 98.5 & 55.4 & 59.8 & 92.6 \\
      & Hybrid Raw    & 88.9 & 55.1 & 93.2 & 47.1 & 49.4 & 92.6 \\
      & Hybrid Final  & 81.4 & 81.1 & 94.0 & 66.9 & 68.4 & 88.3 \\
    \midrule
    \multirow{3}{*}{PrOntoQA}
      & Autoform.     & 100.0 & 100.0 & 100.0 & 100.0 & -- & -- \\
      & Hybrid Raw    & 100.0 & 99.8  & 100.0 & 100.0 & -- & -- \\
      & Hybrid Final  & 100.0 & 100.0 & 100.0 & 100.0 & -- & -- \\
    \bottomrule
  \end{tabular}%
  }
  \caption{Per-label precision/recall (\%) for \textsc{True},
    \textsc{False}, and \textsc{Uncertain}. PrOntoQA has no
    \textsc{Uncertain} label.}
  \label{tab:perlabel}
\end{table}
 
\section{Additional analysis}
\label{app:additional}
 
The main text reports pooled repair results.
Here we include three supplementary diagnostics that are useful for
interpreting them: an instance-level complementarity analysis against
monolithic Autoformalization, per-model repair gains, and per-dataset
prediction distributions.

\subsection{Complementarity with monolithic Autoformalization}
\label{app:complementarity}

Does per-derivation validation add anything beyond the monolithic
\emph{Autoformalization} baseline? Table~\ref{tab:compl} answers by
partitioning instances, pooled over models, on whether the raw verdict
and the monolithic formalisation were each correct. When only the
monolithic formalisation was right, Hybrid recovers $77.7\%$ of those
cases. When only the raw verdict was right, Hybrid
preserves it $95.0\%$ of the time. And in $19.1\%$ of cases where both
were wrong, Hybrid still reaches the gold label, a gain neither
component achieves on its own.

This accounting differs from the repair metrics in
Table~\ref{tab:repair}, which count only instances whose label the
repair stage changes. The \emph{Net} column counts the instances Hybrid
gets right among those where at least one of the two systems was wrong,
minus the instances it gets wrong among those where both were right,
for $+508$ overall. Because it credits Hybrid with every raw-only
success it preserves, it should not be read as a gain over either
system: pooled over models and datasets, Hybrid gets $307$ more
predictions right than Hybrid Raw (Table~\ref{tab:repair}) and $171$
more than Autoformalization.

\begin{table}[h]
  \centering
  \small
  \setlength{\tabcolsep}{3pt}
  \begin{tabular}{l ccc c}
    \toprule
    Dataset & Form.-only & Raw-only & Both-wrong & Net \\
            & recov.\ & preserv.\ & rescued & \\
    \midrule
    FOLIO       & 62.5\% & 100.0\% & 17.5\% & $+163$ \\
    ProofWriter & 91.3\% & 100.0\% & 11.1\% & $+111$ \\
    ProverQA    & 73.9\% & 88.6\%  & 20.9\% & $+233$ \\
    PrOntoQA    & 100.0\% & -- & -- & $+1$ \\
    \midrule
    \textbf{All} & \textbf{77.7\%} & \textbf{95.0\%} & \textbf{19.1\%} & $\mathbf{+508}$ \\
    \bottomrule
  \end{tabular}
  \caption{Benefit/risk complementarity with monolithic Autoformalization, pooled over models. \emph{Form.-only recov.}: of instances only the monolithic formalisation got right, the share Hybrid also gets right. \emph{Raw-only preserv.}: of instances only the raw verdict got right, the share Hybrid preserves. \emph{Both-wrong rescued}: of instances both got wrong, the share Hybrid rescues. \emph{Net}: instances Hybrid gets right where at least one of the two was wrong, minus instances it gets wrong where both were right; distinct from the repair net in Table~\ref{tab:repair}. PrOntoQA has no raw-only or both-wrong instances.}
  \label{tab:compl}
\end{table}

\subsection{Per-model repair gains}
\label{app:permodel}
 
Table~\ref{tab:permodel_repair} breaks down the raw-to-final repair
gain by model and dataset.
Every model improves after repair on FOLIO, ProofWriter and ProverQA.
The largest gains occur where the raw proof-tree stage is weakest,
especially Gemini~3.1~Flash-Lite on ProofWriter and ProverQA, and
GPT-5~Nano on FOLIO and ProverQA.
These are also the pairs where repair fires most often (Table~\ref{tab:permodel_repair}). On PrOntoQA the raw
verdict is already at ceiling, and repair fires once across all models.
After Holm correction, the gain is significant for 8 of the 20
model-dataset pairs; the twelve that are not all have fewer than 20
repair firings.
 
\begin{table*}[h]
  \centering\small
  \setlength{\tabcolsep}{5pt}
  \begin{tabular}{ll rrr rrr r}
    \toprule
    Dataset & Model
      & Raw & Final & $\Delta$
      & Fires & Helpful & Harmful & $p_{\mathrm{Holm}}$ \\
    \midrule
    \multirow{5}{*}{\textbf{FOLIO}}
      & DeepSeek V4 Flash     & 85.2 & 88.5 & $+3.3$  &  6 &  6 & 0 & $.31$ \\
      & GPT-5 Nano            & 74.7 & 87.4 & $+12.6$ & 26 & 24 & 1 & $<.001$ \\
      & Gemini 3.1 Flash-Lite & 82.4 & 89.0 & $+6.6$  & 14 & 12 & 0 & $.007$ \\
      & Grok 4.3              & 84.6 & 88.5 & $+3.8$  &  8 &  7 & 0 & $.17$ \\
      & Qwen 3.5 Flash        & 84.6 & 88.5 & $+3.8$  &  9 &  8 & 1 & $.32$ \\
    \midrule
    \multirow{5}{*}{\textbf{ProofWriter}}
      & DeepSeek V4 Flash     & 94.7 & 96.4 & $+1.8$  &  4 &  4 & 0 & $.75$ \\
      & GPT-5 Nano            & 96.4 & 98.7 & $+2.2$  &  5 &  5 & 0 & $.44$ \\
      & Gemini 3.1 Flash-Lite & 71.1 & 95.1 & $+24.0$ & 58 & 56 & 2 & $<.001$ \\
      & Grok 4.3              & 83.1 & 96.9 & $+13.8$ & 31 & 31 & 0 & $<.001$ \\
      & Qwen 3.5 Flash        & 91.6 & 96.9 & $+5.3$  & 12 & 12 & 0 & $.007$ \\
    \midrule
    \multirow{5}{*}{\textbf{ProverQA}}
      & DeepSeek V4 Flash     & 74.3 & 80.0 & $+5.7$  & 18 & 14 & 2 & $.050$ \\
      & GPT-5 Nano            & 57.6 & 76.2 & $+18.6$ & 55 & 43 & 4 & $<.001$ \\
      & Gemini 3.1 Flash-Lite & 52.9 & 78.1 & $+25.2$ & 66 & 55 & 2 & $<.001$ \\
      & Grok 4.3              & 65.7 & 81.0 & $+15.2$ & 46 & 36 & 4 & $<.001$ \\
      & Qwen 3.5 Flash        & 74.3 & 78.6 & $+4.3$  & 19 & 12 & 3 & $.32$ \\
    \midrule
    \multirow{5}{*}{\textbf{PrOntoQA}}
      & DeepSeek V4 Flash     & 100.0 & 100.0 & $0.0$  & 0 & 0 & 0 & $1.00$ \\
      & GPT-5 Nano            & 100.0 & 100.0 & $0.0$  & 0 & 0 & 0 & $1.00$ \\
      & Gemini 3.1 Flash-Lite & 100.0 & 100.0 & $0.0$  & 0 & 0 & 0 & $1.00$ \\
      & Grok 4.3              & 100.0 & 100.0 & $0.0$  & 0 & 0 & 0 & $1.00$ \\
      & Qwen 3.5 Flash        &  99.5 & 100.0 & $+0.5$ & 1 & 1 & 0 & $1.00$ \\
    \bottomrule
  \end{tabular}
  \caption{Per-model repair breakdown.
    \emph{Raw}/\emph{Final}: accuracy (\%) before and after repair;
    $\Delta$: accuracy gain in points; \emph{Fires}: repair
    invocations; \emph{Helpful}/\emph{Harmful}: firings that move
    toward or away from the gold label; $p_{\mathrm{Holm}}$: exact
    McNemar test of Final against Raw on the Helpful/Harmful counts,
    Holm-corrected within the repair family (Appendix~\ref{app:stats}).}
  \label{tab:permodel_repair}
\end{table*}

\subsection{Per-dataset prediction distributions}
\label{app:prediction}
 
\begin{figure*}[!tp]
  \centering
  \includegraphics[width=0.99\textwidth]{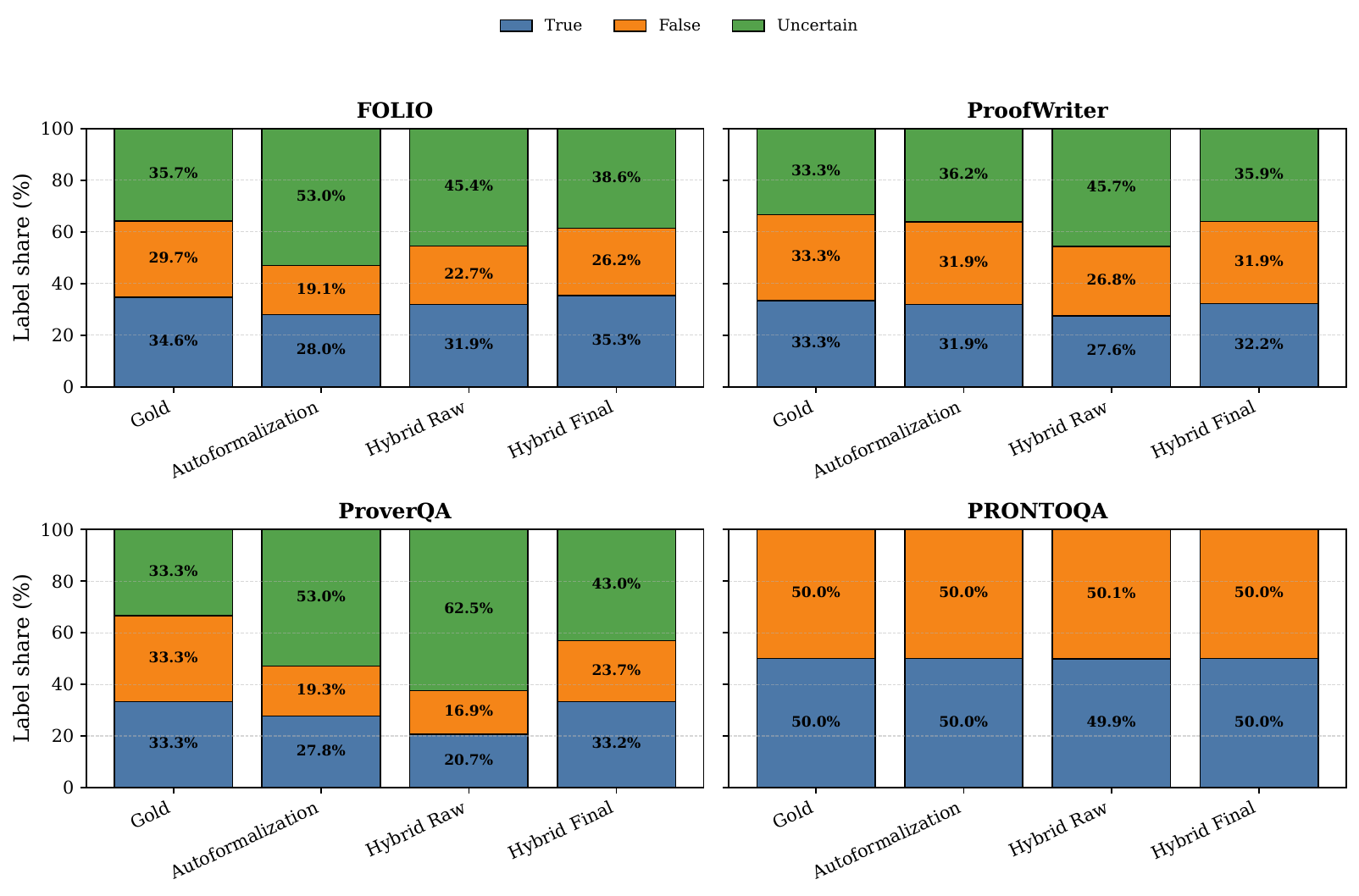}
  \caption{Predicted label distributions for Gold, Autoformalization,
    Hybrid Raw, and Hybrid Final, per dataset.
    Hybrid Raw overpredicts \textsc{Uncertain} on the open-world
    datasets; uncertainty-gated repair shifts mass back toward
    determinate labels and narrows the gap with the gold distribution.}
  \label{fig:label_dist}
\end{figure*}

Figure~\ref{fig:label_dist} shows the label-distribution view of the
repair mechanism.
On the three open-world benchmarks, Hybrid Raw overpredicts
\textsc{Uncertain} relative to the gold distribution, most strongly on
ProverQA ($62.5\%$ predicted \textsc{Uncertain} vs.\ $33.3\%$ gold).
Hybrid Final narrows this gap on every open-world dataset, reducing
the predicted \textsc{Uncertain} share to $38.6\%$ on FOLIO, $35.9\%$
on ProofWriter, and $43.0\%$ on ProverQA.
Since repair revises only Uncertain verdicts, the whole shift consists of resolved abstentions.
 
\section{Statistical testing details}
\label{app:stats}

Each test compares two methods on the same examples with the same model. A method's final prediction is used as in Table~\ref{tab:main}, including its fallback where it has one, and an instance without a prediction counts as incorrect. The tests treat the benchmark examples as the source of variation under the fixed configuration of Appendix~\ref{app:models}; they do not measure run-to-run variation in the providers' sampling.

\paragraph{Per-model tests.} For each dataset, model, and comparison, the exact McNemar test uses only the examples on which the two methods disagree, and asks whether the split between examples only Hybrid gets right and examples only the other method gets right is consistent with a fair coin. We report two-sided exact binomial $p$-values.

\paragraph{Tests of the mean over models.} Each example contributes the average of its five per-model correctness differences, so the five predictions on one example stay together. The $95\%$ CI is the percentile interval of $10{,}000$ bootstrap resamples of the examples. The $p$-value comes from a two-sided sign-flip test: the sign of each example's difference is flipped at random $100{,}000$ times, and $p=(k+1)/100{,}001$, where $k$ is the number of flips whose absolute mean difference is at least the observed one. The smallest attainable value is therefore about $10^{-5}$, and we report values below $.001$ as $p<.001$. All random draws use a fixed seed.

\paragraph{Multiple comparisons.} Holm correction is applied within three families: \emph{primary}, Hybrid against the six baselines on FOLIO and ProverQA ($12$ tests of the mean and $60$ per-model tests); \emph{repair}, Hybrid Final against Hybrid Raw on all four datasets ($4$ and $20$); and \emph{secondary}, Hybrid against the six baselines on ProofWriter and PrOntoQA ($12$ and $60$).

\paragraph{Results.} Table~\ref{tab:sig_all} lists all $28$ tests of the mean and Table~\ref{tab:mcnemar} all $140$ per-model tests. In the primary family, $48$ of the $60$ per-model tests have uncorrected $p<.05$, and $25$ remain significant after Holm correction. Against Logic-LM, Hybrid is ahead for all five models on both FOLIO and ProverQA, but only GPT-5~Nano on FOLIO remains significant after correction. On ProofWriter, no per-model difference against Autoformalization or LINC is significant. On PrOntoQA, two of the $30$ per-model comparisons with baselines are significant: Gemini~3.1~Flash-Lite against Standard ($48$ vs.\ $0$ disagreements) and GPT-5~Nano against Logic-LM ($13$ vs.\ $0$). Against CoT on PrOntoQA, the bootstrap interval excludes zero but the sign-flip test does not reject ($p=.062$ before correction). Only five model-example pairs differ there, all in Hybrid's direction; when no example points the other way, no bootstrap resample can have a negative mean, so we follow the permutation test.

\begin{table}[h]
  \centering\small
  \setlength{\tabcolsep}{4pt}
  \begin{tabular}{@{}c l r c r r@{}}
    \toprule
    & Hybrid vs. & $\Delta$ & 95\% CI & $p$ & $p_{\mathrm{Holm}}$ \\
    \midrule
    \multirow{7}{*}{\rotatebox[origin=c]{90}{\textbf{FOLIO}}}
      & Logic-LM               & $+5.3$  & $[2.3, 8.5]$   & $<.001$ & $<.001$ \\
      & LINC                   & $+10.3$ & $[5.8, 15.1]$  & $<.001$ & $<.001$ \\
      & CLOVER                 & $+12.3$ & $[8.2, 16.6]$  & $<.001$ & $<.001$ \\
      & Standard               & $+8.4$  & $[5.9, 11.0]$  & $<.001$ & $<.001$ \\
      & CoT                    & $+7.5$  & $[5.2, 9.9]$   & $<.001$ & $<.001$ \\
      & Autoform.              & $+10.9$ & $[6.7, 15.2]$  & $<.001$ & $<.001$ \\
      \cmidrule(l){2-6}
      & Hybrid Raw$^{\dagger}$ & $+6.0$  & $[4.0, 8.5]$   & $<.001$ & $<.001$ \\
    \midrule
    \multirow{7}{*}{\rotatebox[origin=c]{90}{\textbf{ProverQA}}}
      & Logic-LM               & $+4.9$  & $[2.4, 7.3]$   & $<.001$ & $<.001$ \\
      & LINC                   & $+14.0$ & $[10.2, 17.7]$ & $<.001$ & $<.001$ \\
      & CLOVER                 & $+6.2$  & $[3.2, 9.1]$   & $<.001$ & $<.001$ \\
      & Standard               & $+13.3$ & $[9.8, 16.8]$  & $<.001$ & $<.001$ \\
      & CoT                    & $+13.7$ & $[10.0, 17.5]$ & $<.001$ & $<.001$ \\
      & Autoform.              & $+7.2$  & $[4.1, 10.5]$  & $<.001$ & $<.001$ \\
      \cmidrule(l){2-6}
      & Hybrid Raw$^{\dagger}$ & $+13.8$ & $[10.4, 17.3]$ & $<.001$ & $<.001$ \\
    \midrule
    \multirow{7}{*}{\rotatebox[origin=c]{90}{\textbf{ProofWriter}}}
      & Logic-LM               & $+15.2$ & $[10.9, 19.7]$ & $<.001$ & $<.001$ \\
      & LINC                   & $+0.4$  & $[-0.6, 1.4]$  & $.62$   & $1.00$ \\
      & CLOVER                 & $+3.2$  & $[2.0, 4.5]$   & $<.001$ & $<.001$ \\
      & Standard               & $+13.6$ & $[11.4, 15.9]$ & $<.001$ & $<.001$ \\
      & CoT                    & $+9.8$  & $[7.4, 12.3]$  & $<.001$ & $<.001$ \\
      & Autoform.              & $-0.4$  & $[-1.2, 0.4]$  & $.51$   & $1.00$ \\
      \cmidrule(l){2-6}
      & Hybrid Raw$^{\dagger}$ & $+9.4$  & $[7.4, 11.6]$  & $<.001$ & $<.001$ \\
    \midrule
    \multirow{7}{*}{\rotatebox[origin=c]{90}{\textbf{PrOntoQA}}}
      & Logic-LM               & $+2.2$  & $[1.3, 3.2]$   & $<.001$ & $<.001$ \\
      & LINC                   & $0.0$   & $[0.0, 0.0]$   & $1.00$  & $1.00$ \\
      & CLOVER                 & $+0.1$  & $[0.0, 0.3]$   & $1.00$  & $1.00$ \\
      & Standard               & $+4.8$  & $[3.7, 6.0]$   & $<.001$ & $<.001$ \\
      & CoT                    & $+0.5$  & $[0.1, 1.0]$   & $.062$  & $.37$ \\
      & Autoform.              & $0.0$   & $[0.0, 0.0]$   & $1.00$  & $1.00$ \\
      \cmidrule(l){2-6}
      & Hybrid Raw$^{\dagger}$ & $+0.1$  & $[0.0, 0.3]$   & $1.00$  & $1.00$ \\
    \bottomrule
  \end{tabular}
  \caption{All tests of Hybrid's accuracy gain averaged over the five models. $\Delta$: Hybrid's accuracy minus the method's (points); \emph{95\% CI}: paired bootstrap interval over examples; $p$: two-sided sign-flip test; $p_{\mathrm{Holm}}$: Holm-corrected within family. Baseline rows under FOLIO and ProverQA form the primary family and those under ProofWriter and PrOntoQA the secondary family; $^{\dagger}$: repair family.}
  \label{tab:sig_all}
\end{table}

\begin{table*}[h]
  \centering\small
  \setlength{\tabcolsep}{4pt}
  \begin{tabular}{ll ccccc}
    \toprule
    & & DeepSeek & GPT-5 & Gemini 3.1 & Grok & Qwen 3.5 \\
    Dataset & Hybrid vs. & V4 Flash & Nano & Flash-Lite & 4.3 & Flash \\
    \midrule
    \multirow{7}{*}{\textbf{FOLIO}}
      & Logic-LM   & $+4.9$ (15/6) & $+9.3^{*}$ (21/4) & $+8.2$ (20/5) & $+2.7$ (13/8) & $+1.1$ (10/8) \\
      & LINC       & $+7.1$ (19/6) & $+9.9^{*}$ (22/4) & $+14.8^{*}$ (32/5) & $+12.6^{*}$ (29/6) & $+7.1$ (24/11) \\
      & CLOVER     & $+13.2^{*}$ (30/6) & $+9.3$ (22/5) & $+13.2^{*}$ (27/3) & $+13.7^{*}$ (32/7) & $+12.1^{*}$ (29/7) \\
      & Standard   & $+3.8$ (9/2) & $+8.2$ (19/4) & $+22.0^{*}$ (41/1) & $+4.4$ (11/3) & $+3.3$ (8/2) \\
      & CoT        & $+6.0$ (13/2) & $+11.0^{*}$ (22/2) & $+12.1^{*}$ (23/1) & $+3.3$ (9/3) & $+4.9$ (12/3) \\
      & Autoform.  & $+9.3$ (23/6) & $+12.6^{*}$ (26/3) & $+11.0^{*}$ (23/3) & $+8.8$ (23/7) & $+12.6^{*}$ (30/7) \\
      & Hybrid Raw & $+3.3$ (6/0) & $+12.6^{*}$ (24/1) & $+6.6^{*}$ (12/0) & $+3.8$ (7/0) & $+3.8$ (8/1) \\
    \midrule
    \multirow{7}{*}{\textbf{ProverQA}}
      & Logic-LM   & $+5.2$ (16/5) & $+8.1$ (27/10) & $+6.2$ (22/9) & $+1.9$ (19/15) & $+2.9$ (16/10) \\
      & LINC       & $+10.0^{*}$ (29/8) & $+12.4^{*}$ (37/11) & $+10.5^{*}$ (30/8) & $+7.1$ (26/11) & $+30.0^{*}$ (71/8) \\
      & CLOVER     & $+9.0^{*}$ (26/7) & $+4.3$ (24/15) & $+8.1$ (27/10) & $+9.0$ (30/11) & $+0.5$ (15/14) \\
      & Standard   & $+9.0$ (28/9) & $+17.1^{*}$ (43/7) & $+29.0^{*}$ (69/8) & $+9.0$ (32/13) & $+2.4$ (16/11) \\
      & CoT        & $+8.6$ (28/10) & $+17.1^{*}$ (42/6) & $+19.0^{*}$ (48/8) & $+17.1^{*}$ (45/9) & $+6.7$ (23/9) \\
      & Autoform.  & $+6.2$ (20/7) & $+6.7$ (28/14) & $+10.0^{*}$ (28/7) & $+6.2$ (23/10) & $+7.1$ (23/8) \\
      & Hybrid Raw & $+5.7$ (14/2) & $+18.6^{*}$ (43/4) & $+25.2^{*}$ (55/2) & $+15.2^{*}$ (36/4) & $+4.3$ (12/3) \\
    \midrule
    \multirow{7}{*}{\textbf{ProofWriter}}
      & Logic-LM   & $+12.9^{*}$ (29/0) & $+20.0^{*}$ (46/1) & $+16.4^{*}$ (44/7) & $+15.6^{*}$ (41/6) & $+11.1^{*}$ (27/2) \\
      & LINC       & $+0.4$ (1/0) & $+1.3$ (4/1) & $-2.2$ (2/7) & $+0.9$ (4/2) & $+1.3$ (5/2) \\
      & CLOVER     & $+1.3$ (3/0) & $+2.7$ (6/0) & $0.0$ (5/5) & $+10.7^{*}$ (26/2) & $+1.3$ (6/3) \\
      & Standard   & $+6.7^{*}$ (15/0) & $+2.2$ (5/0) & $+41.8^{*}$ (97/3) & $+13.3^{*}$ (31/1) & $+4.0$ (10/1) \\
      & CoT        & $+9.3^{*}$ (21/0) & $+2.7$ (7/1) & $+22.2^{*}$ (53/3) & $+11.6^{*}$ (26/0) & $+3.1$ (8/1) \\
      & Autoform.  & $+0.9$ (2/0) & $-0.4$ (0/1) & $+0.4$ (5/4) & $-1.3$ (0/3) & $-1.3$ (1/4) \\
      & Hybrid Raw & $+1.8$ (4/0) & $+2.2$ (5/0) & $+24.0^{*}$ (56/2) & $+13.8^{*}$ (31/0) & $+5.3^{*}$ (12/0) \\
    \midrule
    \multirow{7}{*}{\textbf{PrOntoQA}}
      & Logic-LM   & $+2.0$ (4/0) & $+6.5^{*}$ (13/0) & $+1.0$ (2/0) & $+1.5$ (3/0) & $0.0$ (0/0) \\
      & LINC       & $0.0$ (0/0) & $0.0$ (0/0) & $0.0$ (0/0) & $0.0$ (0/0) & $0.0$ (0/0) \\
      & CLOVER     & $0.0$ (0/0) & $+0.5$ (1/0) & $0.0$ (0/0) & $0.0$ (0/0) & $0.0$ (0/0) \\
      & Standard   & $0.0$ (0/0) & $0.0$ (0/0) & $+24.0^{*}$ (48/0) & $0.0$ (0/0) & $0.0$ (0/0) \\
      & CoT        & $+0.5$ (1/0) & $0.0$ (0/0) & $+2.0$ (4/0) & $0.0$ (0/0) & $0.0$ (0/0) \\
      & Autoform.  & $0.0$ (0/0) & $0.0$ (0/0) & $0.0$ (0/0) & $0.0$ (0/0) & $0.0$ (0/0) \\
      & Hybrid Raw & $0.0$ (0/0) & $0.0$ (0/0) & $0.0$ (0/0) & $0.0$ (0/0) & $+0.5$ (1/0) \\
    \bottomrule
  \end{tabular}
  \caption{Per-model exact McNemar tests. Each cell gives Hybrid's accuracy minus the other method's (points) and, in parentheses, the number of examples only Hybrid gets right / only the other method gets right. $^{*}$: Holm-corrected $p<.05$ within the test's family (Appendix~\ref{app:stats}). The \emph{Hybrid Raw} rows compare Hybrid Final with its own Stage-1 verdict; their counts equal the Helpful and Harmful columns of Table~\ref{tab:permodel_repair}.}
  \label{tab:mcnemar}
\end{table*}

\section{Implementation details}
\label{app:impl}

\paragraph{Structured output and parsing.}
Hybrid Reasoning uses structured model outputs for the proof-tree stage.
Each response is parsed into a goal analysis, an ordered list of
derivations, target matches, and a raw verdict. We reject outputs with
missing required fields, invalid labels, malformed support references,
cyclic dependencies among derivations, or references to derivations that
have not yet been introduced. If the final parsed object remains invalid,
the instance is marked non-executable for Hybrid and falls back to the
CoT fallback prediction.

\paragraph{Retry policy.}
All reported Hybrid experiments use one retry for proof-tree generation:
when the initial structured output fails parsing or structural validation,
we pass the parser error back to the model and ask it to return the same
object in valid format. Hybrid also uses one retry for each
autoformalization subcall: when the initial formalization is unusable, we
rerun that subcall once. The standalone \emph{Autoformalization} baseline
uses the same one-retry policy for its whole-problem formalization call.

\paragraph{Local validation.}
For each derivation $d_i$, the cited support set $S_i$ is converted into a local entailment problem. Premises in $S_i$ may be original input premises or previously generated derivations. The local conclusion is the natural-language claim of $d_i$. This support--claim pair is passed to the autoformalization module, which generates a first-order vocabulary, formalized premises, and a formalized conclusion. A derivation is added to $\Dval$ only if the resulting formal problem is parseable, vocabulary-consistent, and discharged by Prover9. Failed, non-executable, or unproved local checks are omitted from $\Dval$; they are not treated as negative evidence.

\paragraph{Whole-problem autoformalization.}
The monolithic \emph{Autoformalization} baseline uses the same formalization and prover interface as local validation, but applies it once to the full premise set and hypothesis. For open-world datasets, the system separately checks whether $P \vdash h$ and whether $P \vdash \neg h$. If exactly one side is proved, the corresponding determinate label is returned; otherwise the prediction is \textsc{Uncertain}. This gives a direct comparison between full-problem translation and Hybrid's local derivation checks under the same prover interface.

\paragraph{Prover calls.}
All symbolic checks are lowered to Prover9 input and run with a fixed timeout. A prover call is counted as successful only when Prover9 \citep{mccune2010prover9} returns a proof for the queried entailment. Timeout, syntax failure, vocabulary mismatch, or absence of proof are treated as non-validation for local derivations and as non-execution or non-proof for whole-problem formalization, depending on whether the formal problem was usable by the prover.

\paragraph{Repair decision rule.}
Repair is applied only when the raw proof-tree verdict is \textsc{Uncertain}. The repair premise set is $P\cup\Dval$, i.e., the original premises plus only the locally validated derivations. The system checks the hypothesis side first, followed by the negated side if the former is not discharged; when the proof-tree goal analysis decomposes either side into subgoals, those subgoals are checked according to the recorded connective. A single-literal goal requires its only subgoal, a disjunctive goal accepts any proved subgoal, and a conjunctive goal requires all subgoals. If the true side is discharged, the final label is \textsc{True}; if the false side is discharged, it is \textsc{False}; otherwise it remains \textsc{Uncertain}. Raw \textsc{True} and \textsc{False} predictions are preserved and are not overwritten by repair.

\section{Structured output discussion}
\label{app:structured}
Stage~1 could obtain well-formed output in three ways. Grammar-constrained decoding masks every token that would violate a grammar and so guarantees well-formed output \citep{willard, geng}, but it needs control over the model's next-token distribution, which hosted APIs do not give. Several providers instead enforce a JSON schema on their side \citep{openai}. The third option is to prompt for the format and validate the result. We combine the last two: Stage~1 uses the provider's JSON-schema or JSON-object mode where available, then parses each response and re-prompts once with the parser error. Validation is needed even under a strict schema, which cannot express that derivation identifiers are unique or that each support cites an earlier derivation, so we also check the tree's structure after parsing.

Strict output formats can also degrade reasoning, more so the stricter they are, partly because a model may give its answer before its reasoning \citep{tam}. To limit this risk, each goal, derivation, target match and verdict in our schema begins with a free-text reasoning field, so the model reasons before committing to a structured answer. Rather than tightening the format further, we handle malformed output after generation, with the simple corrective step; $34$ of $4{,}085$ responses ($0.8\%$) needed it, and two fell back to CoT. The format does not appear to cost accuracy: Hybrid Raw is within $0.1$ points of CoT on ProverQA and $0.4$--$1.4$ points above it on the other datasets (Figure~\ref{fig:method}), although CoT's output has only two fields.

\section{Ablation study}
\label{app:ablations}

We ablate the evidence used by the uncertainty-gated repair stage. The
main Hybrid system repairs from $P\cup\Dval$, where $\Dval$ contains only
locally validated derivations. We compare it with two variants. The
\emph{all-derivations} variant uses every derivation proposed by the
proof-tree stage, regardless of whether it was validated. The
\emph{no-derivations} variant repairs from the original premise set $P$
alone. All variants reuse the Stage-1 proof trees and raw verdicts of the main
system, so they differ only in Stage~3 and are compared on identical
examples. A second ablation removes the uncertainty gate itself
(\emph{Uncertainty gate}, below).

\begin{table*}[!t]
  \centering\small
  \setlength{\tabcolsep}{4pt}
  \begin{tabular}{ll c ccc ccc}
    \toprule
    & & & \multicolumn{3}{c}{Gated} & \multicolumn{3}{c}{Ungated} \\
    \cmidrule(lr){4-6}\cmidrule(lr){7-9}
    Dataset & Model & Raw & None & All & Validated$^{\dagger}$ & None & All & Validated \\
    \midrule
    \multirow{6}{*}{\textbf{FOLIO}}
      & DeepSeek V4 Flash     & 85.2 & 88.5 & 87.9 & 88.5 & 88.5 & 88.5 & \best{89.0} \\
      & GPT-5 Nano            & 74.7 & 83.0 & 83.5 & \best{87.4} & 83.0 & 83.5 & 86.8 \\
      & Gemini 3.1 Flash-Lite & 82.4 & 87.9 & 80.2 & \best{89.0} & 87.4 & 79.7 & 88.5 \\
      & Grok 4.3              & 84.6 & 87.9 & 87.9 & 88.5 & \best{89.0} & 87.9 & 88.5 \\
      & Qwen 3.5 Flash        & 84.6 & 87.4 & 86.3 & \best{88.5} & 86.8 & 86.3 & \best{88.5} \\
      \cmidrule(l){2-9}
      & \emph{Mean}           & \emph{82.3} & \emph{86.9} & \emph{85.2} & \emph{\best{88.4}} & \emph{86.9} & \emph{85.2} & \emph{88.2} \\
    \midrule
    \multirow{6}{*}{\textbf{ProofWriter}}
      & DeepSeek V4 Flash     & 94.7 & 96.0 & \best{97.3} & 96.4 & 96.0 & \best{97.3} & 96.4 \\
      & GPT-5 Nano            & 96.4 & \best{99.1} & 97.8 & 98.7 & \best{99.1} & 97.8 & 98.7 \\
      & Gemini 3.1 Flash-Lite & 71.1 & \best{96.0} & 81.3 & 95.1 & \best{96.0} & 81.3 & 95.1 \\
      & Grok 4.3              & 83.1 & 97.8 & 96.0 & 96.9 & \best{98.2} & 96.0 & 97.3 \\
      & Qwen 3.5 Flash        & 91.6 & \best{98.2} & 92.9 & 96.9 & \best{98.2} & 92.9 & 96.9 \\
      \cmidrule(l){2-9}
      & \emph{Mean}           & \emph{87.4} & \emph{97.4} & \emph{93.1} & \emph{96.8} & \emph{\best{97.5}} & \emph{93.1} & \emph{96.9} \\
    \midrule
    \multirow{6}{*}{\textbf{ProverQA}}
      & DeepSeek V4 Flash     & 74.3 & 79.5 & 76.7 & \best{80.0} & 79.0 & 76.2 & 79.5 \\
      & GPT-5 Nano            & 57.6 & 73.8 & 75.7 & \best{76.2} & 73.8 & 75.2 & 75.7 \\
      & Gemini 3.1 Flash-Lite & 52.9 & 72.4 & 71.0 & \best{78.1} & 71.9 & 70.5 & \best{78.1} \\
      & Grok 4.3              & 65.7 & 79.5 & \best{82.4} & 81.0 & 79.5 & \best{82.4} & 81.0 \\
      & Qwen 3.5 Flash        & 74.3 & \best{79.5} & 79.0 & 78.6 & 79.0 & 78.1 & 77.6 \\
      \cmidrule(l){2-9}
      & \emph{Mean}           & \emph{65.0} & \emph{77.0} & \emph{77.0} & \emph{\best{78.8}} & \emph{76.7} & \emph{76.5} & \emph{78.4} \\
    \midrule
    \multirow{6}{*}{\textbf{PrOntoQA}}
      & DeepSeek V4 Flash     & 100.0 & \best{100.0} & \best{100.0} & \best{100.0} & \best{100.0} & \best{100.0} & \best{100.0} \\
      & GPT-5 Nano            & 100.0 & \best{100.0} & \best{100.0} & \best{100.0} & \best{100.0} & \best{100.0} & \best{100.0} \\
      & Gemini 3.1 Flash-Lite & 100.0 & \best{100.0} & \best{100.0} & \best{100.0} & \best{100.0} & \best{100.0} & \best{100.0} \\
      & Grok 4.3              & 100.0 & \best{100.0} & \best{100.0} & \best{100.0} & \best{100.0} & \best{100.0} & \best{100.0} \\
      & Qwen 3.5 Flash        & 99.5 & 99.5 & \best{100.0} & \best{100.0} & 99.5 & \best{100.0} & \best{100.0} \\
      \cmidrule(l){2-9}
      & \emph{Mean}           & \emph{99.9} & \emph{99.9} & \emph{\best{100.0}} & \emph{\best{100.0}} & \emph{99.9} & \emph{\best{100.0}} & \emph{\best{100.0}} \\
    \midrule
    \multicolumn{2}{l}{\emph{All datasets}} & \emph{83.6} & \emph{90.3} & \emph{88.8} & \emph{\best{91.0}} & \emph{90.3} & \emph{88.7} & \emph{90.9} \\
    \bottomrule
  \end{tabular}
  \caption{Stage-3 ablations by model: accuracy (\%) of the raw Stage-1
  verdict and of every repair variant. \emph{Gated} repair revises only
  raw \textsc{Uncertain} verdicts; \emph{Ungated} repair runs on every
  instance and overwrites a decisive raw verdict when it discharges the
  opposite side. Columns give the derivations in the repair premise set:
  \emph{None} (repair from $P$ alone), \emph{All} proposed derivations, or
  \emph{Validated} derivations only. $^{\dagger}$The full system.
  \best{Bold}: best of the six variants in each row. \emph{Mean} rows
  average over the five models; the \emph{All datasets} row averages the
  four dataset means.}
  \label{tab:ablation_permodel}
\end{table*}

The ablation isolates the role of validated derivations
(Table~\ref{tab:ablation_permodel}). Using all proposed derivations is not
sufficient. It underperforms the validated system on all three open-world
datasets, and it does no better than repairing without derivations: it
trails that variant on FOLIO and ProofWriter and ties it on ProverQA. This
indicates that intermediate derivations are useful only when they are
filtered; otherwise, unvalidated claims can add noise to the repair
premise set.

The no-derivation variant is a strong ablation because it tests whether
the gain comes merely from rerunning symbolic search on the original
problem when the proof-tree model abstains. It improves substantially
over the raw verdict, especially on ProofWriter and ProverQA, showing
that part of Hybrid's gain comes from giving uncertain instances another
symbolic decision opportunity. However, the full validated-derivation system remains best on FOLIO and ProverQA, the two hardest datasets, and is best on average.

\paragraph{Uncertainty gate.}
The main system repairs only raw \textsc{Uncertain} verdicts. For each
premise set we also run an \emph{ungated} variant, in which repair runs
on every instance: a decisive raw verdict is overwritten when repair
discharges the opposite side, and kept otherwise. The ungated columns of
Table~\ref{tab:ablation_permodel} report the results. Removing the gate
changes no dataset mean by more than $0.5$ points. Across models and
datasets there are $2{,}500$ decisive raw verdicts. With validated
derivations, ungated repair proves the raw label for $2{,}412$ of them,
overturns $13$, and leaves the remaining $75$ unchanged; with no
derivations and with all derivations it overturns $15$ and $10$. In every
setting the overturns that break a correct verdict outnumber those that
fix a wrong one (no derivations: $7$ against $5$; all derivations: $6$
against $1$; validated derivations: $7$ against $3$), so each ungated
variant ends with fewer correct predictions than its gated counterpart.

\paragraph{Repair outcomes.}
Pooled over models and datasets, the relabelling counts explain these
gaps. Under the gate, every harmful relabelling turns a correct
\textsc{Uncertain} verdict into \textsc{True} or \textsc{False}. The full
system makes $19$ of them. With all derivations there are $89$, and in
$72$ of these the repair premise set contains at least one derivation
that failed validation. Repair with all derivations also fixes fewer
wrong verdicts ($306$ against $326$), so its net gain falls from $+307$
to $+217$. Without derivations, repair is slightly more precise than the
full system ($88.3\%$ of relabellings correct, against $86.2\%$) but
relabels fewer instances ($332$ against $378$), for a net gain of
$+281$.

\DeclareRobustCommand{\valid}{{\usefont{U}{pzd}{m}{n}\char51}}
\DeclareRobustCommand{\invalid}{{\usefont{U}{pzd}{m}{n}\char55}}

\section{Error analysis in detail}
\label{app:cases}

All counts pool the five models over the runs reported in Table~\ref{tab:main}. For each case type we drew one instance uniformly at random from the pooled runs, using a single Python \texttt{random.Random(0)} generator in the order A--E, with candidates sorted by dataset, model and instance id; each subsection states the candidate pool. We report every drawn case as it is, including Case~C, whose gold label we believe is wrong. Premises keep the numbering of the model's input; for ProverQA we show only the premises a case relies on. \valid\ marks a derivation accepted by local validation and \invalid\ one that was rejected.

\subsection{Abstentions and where proof chains stop}
\label{app:err-abstain}
Table~\ref{tab:stage1} splits the decidable (gold \textsc{True}/\textsc{False}) instances by the raw verdict and follows the abstentions through repair. On every open-world dataset the raw verdict abstains far more often than it proves the wrong side. On gold-\textsc{Uncertain} instances it claims a proof that does not exist for $8.3\%$ of FOLIO, $0.3\%$ of ProofWriter and $7.4\%$ of ProverQA.

\begin{table}[h]
  \centering\small
  \setlength{\tabcolsep}{3pt}
  \resizebox{\columnwidth}{!}{%
  \begin{tabular}{l r rrr rrr}
    \toprule
    & & \multicolumn{3}{c}{Raw verdict (\%)} & \multicolumn{3}{c}{Abstentions after repair (\%)} \\
    \cmidrule(lr){3-5}\cmidrule(lr){6-8}
    Dataset & $n$ & Correct & Abstain & Opposite & Correct & Wrong & Open \\
    \midrule
    FOLIO       & 585 & 77.1 & 19.7 & 3.1 & 48.7 & 3.5 & 47.8 \\
    ProofWriter & 750 & 81.2 & 18.7 & 0.1 & 77.1 & 0.0 & 22.9 \\
    ProverQA    & 700 & 51.1 & 47.4 & 1.4 & 48.2 & 8.7 & 43.1 \\
    \bottomrule
  \end{tabular}%
  }
  \caption{Stage-1 outcomes on decidable (gold \textsc{True}/\textsc{False}) instances, pooled over models. \emph{Opposite}: the raw verdict proves the wrong side. The right block follows the abstentions through repair: relabelled correctly, relabelled wrongly, or left \textsc{Uncertain} (\emph{Open}). One FOLIO instance has no raw verdict.}
  \label{tab:stage1}
\end{table}

Table~\ref{tab:depth} breaks down ProofWriter's decidable instances by gold proof depth. Trees grow with depth (from $2.4$ to $5.8$ derivations), but beyond depth~1 they more often stop before the goal. The effect is concentrated in two models: DeepSeek~V4~Flash, GPT-5~Nano and Qwen~3.5~Flash abstain on at most $6$ of $30$ instances at any depth, whereas Gemini~3.1~Flash-Lite abstains on up to $20$ and Grok~4.3 on up to $12$. Repair recovers most of these abstentions at every depth. On ProverQA, whose chains are longest, abstaining trees are usually fully validated ($71\%$, against $75\%$ for correct decisive verdicts) but about half as long ($4.9$ against $8.7$ derivations): the model proposes sound steps and stops early. On FOLIO, $26\%$ of abstentions come with an empty proof tree.

\begin{table}[h]
  \centering\small
  \setlength{\tabcolsep}{3pt}
  \resizebox{\columnwidth}{!}{%
  \begin{tabular}{c rrr ccccc}
    \toprule
    & \multicolumn{3}{c}{Pooled} & \multicolumn{5}{c}{Abstentions per model (of 30)} \\
    \cmidrule(lr){2-4}\cmidrule(lr){5-9}
    Depth & Abst. & Final & \#Der. & DS & GPT & Gem. & Grok & Qwen \\
    \midrule
    1 &  2.0 & 100.0 & 2.4 & 0 & 0 &  3 &  0 & 0 \\
    2 & 16.7 &  96.7 & 3.3 & 4 & 2 &  9 &  5 & 5 \\
    3 & 25.3 &  92.7 & 4.1 & 2 & 2 & 20 & 10 & 4 \\
    4 & 24.7 &  89.3 & 5.1 & 6 & 4 & 12 & 10 & 5 \\
    5 & 24.7 &  99.3 & 5.8 & 0 & 0 & 20 & 12 & 5 \\
    \bottomrule
  \end{tabular}%
  }
  \caption{ProofWriter decidable instances by gold proof depth ($n=150$ per depth, pooled over models). \emph{Abst.}: raw abstention rate (\%); \emph{Final}: accuracy after repair (\%); \emph{\#Der.}: mean derivations per tree. Model columns: DeepSeek~V4~Flash, GPT-5~Nano, Gemini~3.1~Flash-Lite, Grok~4.3, Qwen~3.5~Flash.}
  \label{tab:depth}
\end{table}

Cases~A and~B show two chains that stop early. Case~A is drawn from the $39$ ProofWriter abstentions on decidable instances of depth at least~4 with a fully validated tree that repair resolves correctly, Case~B from the $95$ ProverQA abstentions on decidable instances with a fully validated tree that repair leaves unresolved.

\begin{promptbox}[caseinput]{Case A. ProofWriter, depth 5, Grok 4.3 (instance 50)}
(*\textbf{Premises:}*)
P0. Anne is big.
P1. Anne is furry.
P2. Bob is cold.
P3. Bob is red.
P4. Dave is not green.
P5. Gary is not big.
P6. Gary is smart.
P7. If someone is smart and big then they are furry.
P8. Smart people are furry.
P9. Furry, red people are big.
P10. Big people are kind.
P11. If someone is cold then they are smart.
P12. If someone is kind and big then they are not green.
P13. If Gary is cold then Gary is not green.

(*\textbf{Hypothesis:}*)
Bob is green.

(*\textbf{Proof tree:}*)
D1. Bob is smart. (P2, P11)(*~\valid*)
\end{promptbox}

\vspace{0.25cm}

Case A: the chain stops after one step

\textbf{Labels.} Gold \textsc{False}; raw \textsc{Uncertain}; final \textsc{False}.

\textbf{Raw verdict.} ``No premise or derivation connects Bob to being green or not green, so neither the conclusion nor its negation follows.''

\textbf{Diagnosis.} The gold proof has five steps: Bob is smart (P11), furry (P8), big (P9 with P3), kind (P10) and therefore not green (P12). The model proposes the first step correctly and stops, concluding that nothing relates Bob to greenness, although P12 becomes applicable three steps later. The proposed step is sound, but the model stops searching after it. Repair proves \textsc{False}; its formalization entails the negated hypothesis even without D1, so here the prover performed the whole search.

\begin{promptbox}[caseinput]{Case B. ProverQA, GPT-5 Nano (instance 174)}
(*\textbf{Premises (7 of 22):}*)
P2. Forest lives in the rainforest.
P6. Forest does not have venom.
P8. Any snake that has stealth or is agile is adaptable.
P9. Forest is not camouflaged.
P15. If Forest lives in the rainforest, then he is either carnivorous or venomous, but not both.
P18. If Forest is a carnivorous snake, then he is either stealthy or camouflaged, but not both.
P20. If Forest is adaptable, then he can survive in harsh environments and attract scientific attention.

(*\textbf{Hypothesis:}*)
Forest can survive in harsh environments.

(*\textbf{Proof tree:}*)
D1. Forest is carnivorous or venomous, but not both. (P2, P15)(*~\valid*)
D2. Forest is carnivorous. (D1, P6)(*~\valid*)
\end{promptbox}

\vspace{0.25cm}

Case B: the chain stops after two steps, and repair cannot finish it

\textbf{Labels.} Gold \textsc{True}; raw \textsc{Uncertain}; final \textsc{Uncertain}.

\textbf{Raw verdict.} ``The survival claim would require additional assumptions (like adaptability) not provided.''

\textbf{Diagnosis.} Four more steps reach the hypothesis: Forest is stealthy or camouflaged but not both (P18), hence stealthy (P9), hence adaptable (P8), hence able to survive (P20). The model judges adaptability underivable and stops. Its goal analysis is also flawed: the false-side subgoal restates the hypothesis instead of negating it. Repair fails too. Its formalization is not logged when repair fails, but the whole-problem formalization of this instance by the same model and module shows the obstacle: P8 becomes $\forall x\,(\mathit{Snake}(x)\land(\mathit{Stealthy}(x)\lor\mathit{Agile}(x))\to\mathit{Adaptable}(x))$ and P18 is stated for $\mathit{CarnivorousSnake}(\mathit{forest})$, but no premise says that Forest is a snake, because ProverQA's generated text leaves the domain implicit. Adding $\mathit{Snake}(\mathit{forest})$ and $\mathit{Carnivorous}(\mathit{forest})\to\mathit{CarnivorousSnake}(\mathit{forest})$ makes the hypothesis provable; either alone does not.

\subsection{What the prover rejects}
\label{app:err-rejected}
Pooled over the open-world datasets, the prover rejects $2.8\%$ of the steps the model labels modus ponens ($n=4{,}459$), $1.9\%$ of conjunction introductions, $2.5\%$ of conjunction eliminations and $4.0\%$ of disjunctive syllogisms, but $11.5\%$ of universal instantiations, $21.5\%$ of compound steps ($n=967$) and $28.7\%$ of semantic equivalences ($n=265$). FOLIO, with the most naturalistic premises, has the highest rejection rate ($15.0\%$ of derivations, against $5.0\%$ on ProofWriter and $4.7\%$ on ProverQA). Rejected steps fail in three recognisable ways. They rely on knowledge the premises do not state (Cases~C and~D). They compress several inferences and omit a support the step itself uses: the model's own reasoning for the step cites a premise or derivation missing from its support list in $25.5\%$ of rejected compound steps, against $4.2\%$ of accepted ones ($11.6\%$ against $1.7\%$ over all steps). Or they are meta-statements rather than claims, such as ``No information about whether the lion likes the dog is present'' (ProofWriter, Gemini~3.1~Flash-Lite, instance~82).

A derivation is accepted when it follows from its cited supports, even if one of those supports was rejected (Case~C). This holds for $12.8\%$ of accepted derivations on FOLIO, $1.5\%$ on ProofWriter and $4.7\%$ on ProverQA; such derivations were among the repair premises in $30$ of the $378$ relabellings, $25$ of which are correct. Rejections also expose most confident errors. These are rare ($83$ of $3{,}085$ open-world predictions), but $58$ of them rest on a target match that failed validation or was never made. The gate keeps them all, since it never overrides a decisive verdict; this is the signal a less conservative policy could use.

Case~C is drawn from the $31$ FOLIO abstentions on decidable instances with at least one rejected derivation, Case~D from the $28$ confidently wrong raw verdicts on FOLIO or ProverQA whose target match failed validation.

\begin{promptbox}[caseinput]{Case C. FOLIO, DeepSeek V4 Flash (instance 78)}
(*\textbf{Premises:}*)
P0. If the restaurant is listed in Yelp's recommendations, then the restaurant does not receive many negative reviews.
P1. All restaurants with a rating greater than 9 are listed in Yelp's recommendations.
P2. Some restaurants that do not provide take-out service receive many negative reviews.
P3. All restaurants that are popular among local residents have ratings greater than 9.
P4. Subway has a rating greater than 9 or is popular among local residents.

(*\textbf{Hypothesis:}*)
Subway does not provide take-out service.

(*\textbf{Proof tree:}*)
D1. Subway has a rating greater than 9 or is popular among local residents. (P4)(*~\valid*)
D2. If Subway is popular among local residents, then Subway has a rating greater than 9. (P3)(*~\invalid*)
D3. If Subway has a rating greater than 9, then Subway is listed in Yelp's recommendations. (P1)(*~\invalid*)
D4. If Subway is listed in Yelp's recommendations, then Subway does not receive many negative reviews. (P0)(*~\valid*)
D5. Subway has a rating greater than 9. (D1,(*~*)D2)(*~\valid*)
D6. Subway is listed in Yelp's recommendations. (D5, D3)(*~\valid*)
D7. Subway does not receive many negative reviews. (D6, D4)(*~\valid*)
\end{promptbox}

\vspace{0.25cm}

Case C: unstated knowledge is rejected, and the gold label is doubtful

\textbf{Labels.} Gold \textsc{False}; raw \textsc{Uncertain}; final \textsc{Uncertain}.

\textbf{Raw verdict.} ``We can derive that Subway does not receive many negative reviews, but no connection to take-out service. P2 only says some such restaurants exist, not about Subway.''

\textbf{Diagnosis.} The two rejected steps instantiate the premises that quantify over restaurants (P1, P3), and no premise states that Subway is a restaurant: the model supplies this world knowledge silently, and the prover does not accept it. D4, whose premise speaks of ``the restaurant'' generically, is accepted. Because validation is local, D5--D7 are accepted relative to their cited supports even though they rest on D2 and D3. The verdict itself is consistent with the premises: since P2 is existential, neither the hypothesis nor its negation follows, with or without ``Subway is a restaurant'' (checked with Z3). The gold label \textsc{False} follows only if P2 is read as universal, so we believe the gold label is wrong here.

\begin{promptbox}[caseinput]{Case D. FOLIO, Grok 4.3 (instance 20)}
(*\textbf{Premises:}*)
P0. Beijing is the capital of the People's Republic of China.
P1. Beijing is the world's most populous national capital city.
P2. Beijing is located in Northern China.
P3. Beijing hosted the 2008 Summer Olympics and 2008 Summer Paralympics Games.
P4. Beijing has hosted both the Summer and Winter Olympics, along with the Summer and Winter Paralympics
P5. Many of Beijing's 91 universities consistently rank among the best in the Asia-Pacific and the world.

(*\textbf{Hypothesis:}*)
Beijing is located in southern China.

(*\textbf{Proof tree:}*)
D1. Beijing is located in Northern China. (P2)(*~\valid*)
D2. Beijing is not located in southern China. (D1; semantic equivalence)(*~\invalid*)
Matches of D2 to the false goal: blocked
\end{promptbox}

\vspace{0.25cm}

Case D: a rejected step behind a confident verdict

\textbf{Labels.} Gold \textsc{Uncertain}; raw \textsc{False}; final \textsc{False}.

\textbf{Raw verdict.} ``P2 directly entails the negation of the conclusion via location exclusivity.''

\textbf{Diagnosis.} The model treats northern and southern China as mutually exclusive, which the premises do not state. The prover rejects D2, so the match to the false goal is blocked and nothing validated supports the verdict. Because the raw verdict is decisive, the gate keeps it and the error stands.

\subsection{How repair fails}
\label{app:err-repair}
The repair prover made $378$ relabellings (FOLIO $62$, ProofWriter $110$, ProverQA $203$). Of the $378$ relabellings, $52$ are wrong. We checked each relabelling's logged formalization with Z3~\citep{z3}. In $41$ of the $52$ wrong ones ($39$ on ProverQA, $2$ on ProofWriter), the formalized premise set $P\cup\Dval$ is unsatisfiable, so every goal is provable; because the hypothesis is tried before its negation, all $41$ become \textsc{True} ($28$ gold-\textsc{False}, $13$ gold-\textsc{Uncertain}). This accounts for $28$ of the $33$ gold-\textsc{False} instances that repair moves to \textsc{True} in Figure~\ref{fig:confusion}. A further $28$ correct relabellings also rest on unsatisfiable premise sets; they are right only because the gold label happened to be \textsc{True}.

The contradictions have two sources. In $56$ of the $69$ vacuous relabellings, all on ProverQA, the formalized original premises are already inconsistent; in $17$ of them an ``either \ldots\ or'' premise was formalized inclusively although ProverQA's generator uses it exclusively (Case~E). In the other $13$, a validated derivation introduces the contradiction; several are meta-statements of the kind described above that passed validation, such as ``Cannot conclude Charlie is nice'', formalized as a negated tautology. The remaining $10$ wrong relabellings, including all $6$ on FOLIO, come from consistent formalizations, so the error lies in what the formulas or subgoals say rather than in a contradiction. Refusing to relabel when $P\cup\Dval$ is unsatisfiable would raise ProverQA's repair precision to $96.4\%$ ($133$ of $138$) but lower its accuracy by $1.5$ points, since $27$ of its vacuous relabellings matched the gold label.

Case~E is drawn from the $29$ ProverQA abstentions on decidable instances that repair resolves to the wrong label.

\begin{promptbox}[caseinput]{Case E. ProverQA, GPT-5 Nano (instance 50)}
(*\textbf{Premises (10 of 35):}*)
P4. Michelle embraces uncertainty.
P5. If Michelle either embraces uncertainty or thinks outside the box (but not both), then she is able to find unique solutions.
P7. If Michelle pursues her passions, then she values diversity and is open-minded.
P9. Michelle is curious or explores new ideas.
P16. Everyone who either thinks outside the box or brings joy to adults is able to express themselves.
P20. If Michelle is open-minded, then she is either curious or able to find unique solutions (but not both).
P22. If Michelle is creative, then she can either inspire children or bring joy to adults, but not both.
P28. Michelle does not explore new ideas.
P30. Michelle is not able to express herself.
P34. Michelle pursues her passions.

(*\textbf{Hypothesis:}*)
Michelle is creative.

(*\textbf{Proof tree:}*)
D1. Michelle pursues her passions. (P34)(*~\valid*)
D2. Michelle values diversity. (P7, D1)(*~\valid*)
D3. Michelle is open-minded. (P7, D1)(*~\valid*)
\end{promptbox}

\vspace{0.25cm}

Case E: a vacuous repair from an inconsistent formalization

\textbf{Labels.} Gold \textsc{False}; raw \textsc{Uncertain}; final \textsc{True}.

\textbf{Raw verdict.} ``The premises yield that Michelle pursues her passions and from that derive values diversity and openness, but they do not entail that Michelle is creative or not creative.''

\textbf{Diagnosis.} The model stops after three steps. Repair then formalizes P16 with an inclusive disjunction, $\forall x\,((\mathit{ThinksOutsideBox}(x)\lor\mathit{BringsJoyToAdults}(x))\to\mathit{ExpressSelf}(x))$. Michelle thinks outside the box (from D3, P20, P9, P28, P5 and P4), so this formula forces $\mathit{ExpressSelf}(\mathit{michelle})$, contradicting P30. The formalized premise set is unsatisfiable, the prover proves the first goal it tries, and the label becomes \textsc{True}. Read exclusively, as ProverQA's generator intends, the premises are consistent and entail that Michelle is not creative, the gold label: she brings joy to adults (P16, P30) and inspires children (P2, P11, P12, P15, P18, P21, P31; not shown), which P22 rules out for a creative person.

\section{Prompts and few-shot examples}
\label{app:prompts}

We report the prompt templates and one-shot demonstrations used by the
internal methods. Premises are serialized as \texttt{P0}, \texttt{P1},
\ldots, and the queried hypothesis is serialized as \texttt{C0}. The
Standard, CoT, and whole-problem Autoformalization baselines use the
shared one-shot demonstration shown below. Hybrid Reasoning uses one dataset-specific one-shot example for each
dataset path. For FOLIO, ProofWriter, and ProverQA, the one-shot examples
are drawn from publicly available non-evaluation splits of the
corresponding datasets. For PrOntoQA, the one-shot example is drawn from
the left-out portion of the source split. None of the one-shot examples
appears in the corresponding evaluation split.

\subsection{Baseline prompts}

\begin{promptbox}{Standard baseline prompt}
You solve natural-language logical reasoning problems.

You are given:
- A list of premise statements.
- One question statement.

Your task is to decide whether the question is entailed by the premises.

Labels:
- "True": the premises entail the question.
- "False": the premises entail the negation of the question.
- "Uncertain": the premises entail neither the question nor its negation.

Use only the given premises. Do not use outside knowledge except for ordinary language understanding.

Output valid JSON only using exactly this schema:

{
  "prediction": "True | False | Uncertain"
}
\end{promptbox}

\begin{promptbox}{Standard baseline one-shot user message}
Problem:
P0. Lawton Park is a neighbourhood in Seattle.
P1. All citizens of Lawton Park use the zip code 98199.
P2. Tom is a citizen of Lawton Park.

Question:
C0. Tom uses the zip code 98199.
###
\end{promptbox}

\begin{promptbox}{Standard baseline one-shot assistant message}
{
  "prediction": "True"
}
\end{promptbox}

\begin{promptbox}{CoT baseline prompt}
You solve natural-language logical reasoning problems.

You are given:
- A list of premise statements.
- One question statement.

Your task is to decide whether the question is entailed by the premises.

Labels:
- "True": the premises entail the question.
- "False": the premises entail the negation of the question.
- "Uncertain": the premises entail neither the question nor its negation.

Use only the given premises. Do not use outside knowledge except for ordinary language understanding.

First write a short reasoning explaining how the premises support or fail to support the question.
Then write the prediction.

Output valid JSON only using exactly this schema:

{
  "reasoning": "string",
  "prediction": "True | False | Uncertain"
}
\end{promptbox}

\begin{promptbox}{CoT baseline one-shot user message}
Problem:
P0. Lawton Park is a neighbourhood in Seattle.
P1. All citizens of Lawton Park use the zip code 98199.
P2. Tom is a citizen of Lawton Park.

Question:
C0. Tom uses the zip code 98199.
###
\end{promptbox}

\begin{promptbox}{CoT baseline one-shot assistant message}
{
  "reasoning": "P2 states that Tom is a citizen of Lawton Park. P1 states that all citizens of Lawton Park use the zip code 98199. Therefore, Tom uses the zip code 98199.",
  "prediction": "True"
}
\end{promptbox}

\begin{promptbox}{Whole-problem autoformalization prompt}
You translate a natural-language reasoning problem into first-order logic.

You are given:
- One full problem consisting of premise statements.
- One or more conclusion statements to formalize against the same problem.

Your task is to output:
1. Predicates/constants used in the formalization.
2. Formalized premises.
3. Formalized conclusions.

Use only the predicates and constants you define in the output. Do not invent extra symbols in formulas.

The grammar of the first-order logic formula is defined as follows:

1) logical conjunction of expr1 and expr2: expr1 ∧ expr2
2) logical disjunction of expr1 and expr2: expr1 ∨ expr2
3) logical exclusive disjunction of expr1 and expr2: expr1 ⊕ expr2
4) logical negation of expr1: ¬expr1
5) expr1 implies expr2: expr1 → expr2
6) expr1 if and only if expr2: expr1 ↔ expr2
7) logical universal quantification: ∀x
8) logical existential quantification: ∃x

Symbol rules:
- Predicate names must use PascalCase.
- Constants must use lowercase or snake_case.
- Every predicate used in a formula must appear in "predicates".
- Every constant used in a formula must appear in "constants".
- Predicate arity must match its use in formulas.
- Use the same symbol consistently across premises and conclusions.
- Formalize every premise statement and every conclusion statement exactly once.
- Premise IDs must match the numbered premise statements you are given: "P0", "P1", ..., "Pn".
- Conclusion IDs must match the numbered conclusion statements you are given: "C0", "C1", ..., "Ck".
- Do not invent extra output fields.
- Do not include natural-language explanations outside JSON.
- Do not add explanations, commentary, or markdown.

Output exactly one JSON object with this schema:

{
  "predicates": [
    {
      "name": "PredicateName",
      "arity": 1,
      "description": "x is ..."
    }
  ],
  "constants": [
    {
      "name": "constant_name",
      "description": "..."
    }
  ],
  "premises": [
    {
      "id": "P0",
      "formula": "..."
    }
  ],
  "conclusions": [
    {
      "id": "C0",
      "formula": "..."
    }
  ]
}

Output JSON only.
\end{promptbox}

\begin{promptbox}{Whole-problem autoformalization one-shot user message}
Problem:
P0. Lawton Park is a neighbourhood in Seattle.
P1. All citizens of Lawton Park use the zip code 98199.
P2. Tom is a citizen of Lawton Park.

Question:
C0. Tom uses the zip code 98199.
C1. Daniel uses the zip code 98199.
###
\end{promptbox}

\begin{promptbox}{Whole-problem autoformalization one-shot assistant message}
{
  "predicates": [
    {
      "name": "NeighbourhoodIn",
      "arity": 2,
      "description": "x is a neighbourhood in y"
    },
    {
      "name": "CitizenOf",
      "arity": 2,
      "description": "x is a citizen of y"
    },
    {
      "name": "UsesZipCode",
      "arity": 2,
      "description": "x uses zip code y"
    }
  ],
  "constants": [
    {
      "name": "lawton_park",
      "description": "Lawton Park"
    },
    {
      "name": "seattle",
      "description": "Seattle"
    },
    {
      "name": "tom",
      "description": "Tom"
    },
    {
      "name": "daniel",
      "description": "Daniel"
    },
    {
      "name": "zip_98199",
      "description": "the zip code 98199"
    }
  ],
  "premises": [
    {
      "id": "P0",
      "formula": "NeighbourhoodIn(lawton_park, seattle)"
    },
    {
      "id": "P1",
      "formula": "∀x (CitizenOf(x, lawton_park) → UsesZipCode(x, zip_98199))"
    },
    {
      "id": "P2",
      "formula": "CitizenOf(tom, lawton_park)"
    }
  ],
  "conclusions": [
    {
      "id": "C0",
      "formula": "UsesZipCode(tom, zip_98199)"
    },
    {
      "id": "C1",
      "formula": "UsesZipCode(daniel, zip_98199)"
    }
  ]
}
\end{promptbox}

\subsection{Hybrid prompts}

The Hybrid proof-tree prompt is used for Stage~1. Local derivation
validation and target-side repair use the same autoformalization
interface as the whole-problem Autoformalization baseline, but with the
local supports as premises and the derivation claim or target statement
as the conclusion.

\begin{promptbox}{Hybrid proof-tree prompt}
You solve a natural-language logical reasoning problem.

You are given:
- A list of premise statements.
- One conclusion statement.

Your task is to produce a compact proof tree and predict whether the conclusion is entailed by the premises.

Labels:
- "True": the premises entail the conclusion.
- "False": the premises entail the negation of the conclusion.
- "Uncertain": neither the conclusion nor its negation is entailed.

Goal rules:
- For an atomic conclusion A:
  True goal is A.
  False goal is not A.
- For a negated conclusion not A:
  True goal is not A.
  False goal is A.
- For a conjunction A and B:
  True goal requires both A and B.
  False goal requires not A or not B.
- For a disjunction A or B:
  True goal requires A or B.
  False goal requires both not A and not B.
- For an implication if A then B:
  Treat the true goal as not A or B.
  Treat the false goal as A and not B.
- For a biconditional A iff B:
  True goal requires both directions.
  False goal requires one side true and the other side false.
- For a universal conclusion:
  True goal is the universal statement.
  False goal is a counterexample.
- For an existential conclusion:
  True goal is a witness or the existential statement.
  False goal is that no such object exists.

The field composition controls how subgoals are interpreted:
- "single": prove the statement itself or the single main subgoal.
- "any": proving any subgoal satisfies the goal.
- "all": every subgoal must be proved to satisfy the goal.

Important:
- Output JSON only.
- Do not output formal logic.
- Derivation claims must be natural-language statements, not formulas.
- Even if you cannot fully prove the conclusion, generate useful intermediate derivations from the premises whenever possible.

Output exactly one JSON object with this schema:

{
  "goal_analysis": {
    "main_connective": "atomic | negation | conjunction | disjunction | implication_as_disjunction | biconditional | universal | existential | complex",
    "true_goal": {
      "reasoning": "string",
      "composition": "single | any | all",
      "statement": "string",
      "subgoals": [
        {
          "reasoning": "string",
          "id": "T1",
          "role": "string",
          "statement": "string"
        }
      ]
    },
    "false_goal": {
      "reasoning": "string",
      "composition": "single | any | all",
      "statement": "string",
      "subgoals": [
        {
          "reasoning": "string",
          "id": "F1",
          "role": "string",
          "statement": "string"
        }
      ]
    }
  },
  "proof_tree": {
    "derivations": [
      {
        "reasoning": "string",
        "id": "D1",
        "claim": "string",
        "support": ["P0", "D1"],
        "rule": "direct_premise | universal_instantiation | existential_instantiation | existential_introduction | modus_ponens | modus_tollens | conjunction_elimination | conjunction_introduction | disjunction_introduction | disjunctive_syllogism | hypothetical_syllogism | biconditional_elimination | contradiction | counterexample | semantic_equivalence | compound"
      }
    ],
    "target_matches": [
      {
        "reasoning": "string",
        "target": "T | F | T1 | F1",
        "derivations": ["D1"],
        "match_type": "exact | equivalent | entails"
      }
    ],
    "verdict": {
      "reasoning": "string",
      "label": "True | False | Uncertain",
      "satisfied_target": "T | F | none"
    }
  }
}

Rules:
- Use [] for derivations if no useful intermediate statement can be derived.
- Prefer generating useful intermediate derivations over leaving the proof tree empty.
- Use [] for target_matches if no derivation clearly matches a proof goal.
- A derivation may only support later derivations, never earlier ones.
- support entries must be IDs such as P0, P1, D1, D2.
- target_matches should point to derivations that satisfy the target or subgoal.
- target_matches.target must be one of T, F, T1, T2, ..., F1, F2, ...
- Every subgoal statement must be a standalone proposition, not an explanation.
- Put reasoning first inside each object and put explanations there, not inside statement fields.
\end{promptbox}

\subsection{FOLIO Hybrid one-shot demonstration}

\begin{promptbox}{FOLIO Hybrid one-shot user message}
Problem:
P0. An animal is either a cat or a dog.
P1. All dogs bark.
P2. Ghosts do not exist.
P3. If something barks, then it is not alive.
P4. A thing is either alive or dead.
P5. Jojo is not dead.

Question:
C0. If Jojo is a cat or a ghost, then Jojo either barks or is a dog.
###
\end{promptbox}

\begin{promptbox}{FOLIO Hybrid one-shot assistant message}
{
  "goal_analysis": {
    "main_connective": "implication_as_disjunction",
    "true_goal": {
      "reasoning": "The conclusion is an implication A -> B. In this schema, the true goal is represented as not A or B.",
      "composition": "any",
      "statement": "Jojo is not a cat and Jojo is not a ghost, or Jojo either barks or is a dog.",
      "subgoals": [
        {
          "reasoning": "This proves the implication true by making the antecedent false.",
          "id": "T1",
          "role": "negated_antecedent",
          "statement": "Jojo is not a cat and Jojo is not a ghost."
        },
        {
          "reasoning": "This proves the implication true by deriving the consequent.",
          "id": "T2",
          "role": "consequent",
          "statement": "Jojo either barks or is a dog."
        }
      ]
    },
    "false_goal": {
      "reasoning": "An implication A -> B is false when A is true and B is false.",
      "composition": "all",
      "statement": "Jojo is a cat or a ghost, and Jojo neither barks nor is a dog.",
      "subgoals": [
        {
          "reasoning": "This makes the antecedent true.",
          "id": "F1",
          "role": "antecedent",
          "statement": "Jojo is a cat or a ghost."
        },
        {
          "reasoning": "This makes the consequent false.",
          "id": "F2",
          "role": "negated_consequent",
          "statement": "Jojo does not bark and Jojo is not a dog."
        }
      ]
    }
  },
  "proof_tree": {
    "derivations": [
      {
        "reasoning": "P5 says Jojo is not dead. P4 says a thing is either alive or dead, so Jojo is alive.",
        "id": "D1",
        "claim": "Jojo is alive.",
        "support": ["P4", "P5"],
        "rule": "disjunctive_syllogism"
      },
      {
        "reasoning": "P3 says that if something barks, then it is not alive. Since Jojo is alive, Jojo does not bark.",
        "id": "D2",
        "claim": "Jojo does not bark.",
        "support": ["P3", "D1"],
        "rule": "modus_tollens"
      },
      {
        "reasoning": "P1 says all dogs bark. Since Jojo does not bark, Jojo is not a dog.",
        "id": "D3",
        "claim": "Jojo is not a dog.",
        "support": ["P1", "D2"],
        "rule": "modus_tollens"
      },
      {
        "reasoning": "P0 says an animal is either a cat or a dog. Since Jojo is not a dog, Jojo is a cat.",
        "id": "D4",
        "claim": "Jojo is a cat.",
        "support": ["P0", "D3"],
        "rule": "disjunctive_syllogism"
      },
      {
        "reasoning": "If Jojo is a cat, then Jojo is a cat or a ghost.",
        "id": "D5",
        "claim": "Jojo is a cat or a ghost.",
        "support": ["D4"],
        "rule": "disjunction_introduction"
      },
      {
        "reasoning": "Combining Jojo does not bark and Jojo is not a dog gives the negation of the consequent.",
        "id": "D6",
        "claim": "Jojo does not bark and Jojo is not a dog.",
        "support": ["D2", "D3"],
        "rule": "conjunction_introduction"
      }
    ],
    "target_matches": [
      {
        "reasoning": "D5 satisfies the antecedent subgoal.",
        "target": "F1",
        "derivations": ["D5"],
        "match_type": "exact"
      },
      {
        "reasoning": "D6 satisfies the negated consequent subgoal.",
        "target": "F2",
        "derivations": ["D6"],
        "match_type": "exact"
      },
      {
        "reasoning": "F1 and F2 together satisfy the false goal.",
        "target": "F",
        "derivations": ["D5", "D6"],
        "match_type": "entails"
      }
    ],
    "verdict": {
      "reasoning": "The antecedent is derivable and the consequent is refuted, so the implication is false.",
      "label": "False",
      "satisfied_target": "F"
    }
  }
}
\end{promptbox}

\subsection{ProofWriter Hybrid one-shot demonstration}

\begin{promptbox}{ProofWriter Hybrid one-shot user message}
Problem:
P0. Erin is smart.
P1. If something is furry then it is not blue.
P2. Nice things are not cold.
P3. Furry things are cold.
P4. White things are cold.
P5. If something is cold then it is not white.
P6. Smart things are furry.
P7. If something is blue then it is quiet.

Question:
C0. Erin is not white.
###
\end{promptbox}

\begin{promptbox}{ProofWriter Hybrid one-shot assistant message}
{
  "goal_analysis": {
    "main_connective": "negation",
    "true_goal": {
      "reasoning": "The conclusion is a negated atomic statement, so the true goal is to derive that Erin is not white.",
      "composition": "single",
      "statement": "Erin is not white.",
      "subgoals": [
        {
          "reasoning": "This is the negated atomic conclusion itself.",
          "id": "T1",
          "role": "main_goal",
          "statement": "Erin is not white."
        }
      ]
    },
    "false_goal": {
      "reasoning": "To make a negated atomic conclusion false, the corresponding positive atomic statement must be true.",
      "composition": "single",
      "statement": "Erin is white.",
      "subgoals": [
        {
          "reasoning": "This is the positive form of the conclusion.",
          "id": "F1",
          "role": "positive_goal",
          "statement": "Erin is white."
        }
      ]
    }
  },
  "proof_tree": {
    "derivations": [
      {
        "reasoning": "P0 directly states that Erin is smart.",
        "id": "D1",
        "claim": "Erin is smart.",
        "support": ["P0"],
        "rule": "direct_premise"
      },
      {
        "reasoning": "P6 says smart things are furry. Since Erin is smart, Erin is furry.",
        "id": "D2",
        "claim": "Erin is furry.",
        "support": ["P6", "D1"],
        "rule": "modus_ponens"
      },
      {
        "reasoning": "P3 says furry things are cold. Since Erin is furry, Erin is cold.",
        "id": "D3",
        "claim": "Erin is cold.",
        "support": ["P3", "D2"],
        "rule": "modus_ponens"
      },
      {
        "reasoning": "P5 says that if something is cold then it is not white. Since Erin is cold, Erin is not white.",
        "id": "D4",
        "claim": "Erin is not white.",
        "support": ["P5", "D3"],
        "rule": "modus_ponens"
      }
    ],
    "target_matches": [
      {
        "reasoning": "D4 directly matches the true goal subgoal.",
        "target": "T1",
        "derivations": ["D4"],
        "match_type": "exact"
      },
      {
        "reasoning": "D4 directly proves the true goal.",
        "target": "T",
        "derivations": ["D4"],
        "match_type": "exact"
      }
    ],
    "verdict": {
      "reasoning": "The derivation chain proves Erin is not white, so the negated atomic conclusion is true.",
      "label": "True",
      "satisfied_target": "T"
    }
  }
}
\end{promptbox}

\subsection{PrOntoQA Hybrid one-shot demonstration}

\begin{promptbox}{PrOntoQA Hybrid one-shot user message}
Problem:
P0. Jompuses are not shy.
P1. Jompuses are yumpuses.
P2. Each yumpus is aggressive.
P3. Each yumpus is a dumpus.
P4. Dumpuses are not wooden.
P5. Dumpuses are wumpuses.
P6. Wumpuses are red.
P7. Every wumpus is an impus.
P8. Each impus is opaque.
P9. Impuses are tumpuses.
P10. Numpuses are sour.
P11. Tumpuses are not sour.
P12. Tumpuses are vumpuses.
P13. Vumpuses are earthy.
P14. Every vumpus is a zumpus.
P15. Zumpuses are small.
P16. Zumpuses are rompuses.
P17. Max is a yumpus.

Question:
C0. Max is sour.
###
\end{promptbox}

\begin{promptbox}{PrOntoQA Hybrid one-shot assistant message}
{
  "goal_analysis": {
    "main_connective": "atomic",
    "true_goal": {
      "reasoning": "The conclusion is an atomic statement, so the true goal is to derive that Max is sour.",
      "composition": "single",
      "statement": "Max is sour.",
      "subgoals": [
        {
          "reasoning": "This is the atomic conclusion itself.",
          "id": "T1",
          "role": "main_goal",
          "statement": "Max is sour."
        }
      ]
    },
    "false_goal": {
      "reasoning": "For an atomic conclusion, the false goal is the negated atomic statement.",
      "composition": "single",
      "statement": "Max is not sour.",
      "subgoals": [
        {
          "reasoning": "This is the negation of the atomic conclusion.",
          "id": "F1",
          "role": "negated_goal",
          "statement": "Max is not sour."
        }
      ]
    }
  },
  "proof_tree": {
    "derivations": [
      {
        "reasoning": "P17 directly states that Max is a yumpus.",
        "id": "D1",
        "claim": "Max is a yumpus.",
        "support": ["P17"],
        "rule": "direct_premise"
      },
      {
        "reasoning": "P3 says each yumpus is a dumpus. Since Max is a yumpus, Max is a dumpus.",
        "id": "D2",
        "claim": "Max is a dumpus.",
        "support": ["P3", "D1"],
        "rule": "modus_ponens"
      },
      {
        "reasoning": "P5 says dumpuses are wumpuses. Since Max is a dumpus, Max is a wumpus.",
        "id": "D3",
        "claim": "Max is a wumpus.",
        "support": ["P5", "D2"],
        "rule": "modus_ponens"
      },
      {
        "reasoning": "P7 says every wumpus is an impus. Since Max is a wumpus, Max is an impus.",
        "id": "D4",
        "claim": "Max is an impus.",
        "support": ["P7", "D3"],
        "rule": "modus_ponens"
      },
      {
        "reasoning": "P9 says impuses are tumpuses. Since Max is an impus, Max is a tumpus.",
        "id": "D5",
        "claim": "Max is a tumpus.",
        "support": ["P9", "D4"],
        "rule": "modus_ponens"
      },
      {
        "reasoning": "P11 says tumpuses are not sour. Since Max is a tumpus, Max is not sour.",
        "id": "D6",
        "claim": "Max is not sour.",
        "support": ["P11", "D5"],
        "rule": "modus_ponens"
      }
    ],
    "target_matches": [
      {
        "reasoning": "D6 directly matches the false-goal subgoal.",
        "target": "F1",
        "derivations": ["D6"],
        "match_type": "exact"
      },
      {
        "reasoning": "D6 directly proves the false goal.",
        "target": "F",
        "derivations": ["D6"],
        "match_type": "exact"
      }
    ],
    "verdict": {
      "reasoning": "The premises derive the negation of the conclusion, so the conclusion is false.",
      "label": "False",
      "satisfied_target": "F"
    }
  }
}
\end{promptbox}

\subsection{ProverQA Hybrid one-shot demonstration}

\begin{promptbox}{ProverQA Hybrid one-shot user message}
Problem:
P0. Reece earns recognition.
P1. If Simon stays updated on his field or networks with peers, then he will attend conferences.
P2. Simon reads medical literature.
P3. Rhys teaches medical students.
P4. For Rhys, advancing the medical field and teaching medical students are mutually exclusive pursuits.
P5. If Simon has expertise, then he will publish papers.
P6. If Simon has curiosity, then he will pursue medical knowledge.
P7. Simon earns recognition.
P8. If someone earns recognition in their field, then they have expertise.
P9. Milan reads medical literature.
P10. If Milan publishes papers or collaborates with peers, then he attends conferences.
P11. If Milan has expertise, then he will publish papers.
P12. If Reece dedicates himself to research, then he can either make groundbreaking discoveries or improve patient care, but not both.
P13. If Simon publishes papers or collaborates with peers, then he attends conferences.
P14. Anyone who reads medical literature or attends conferences pursues medical knowledge.
P15. Anyone who conducts experiments and collects data can develop expertise.

Question:
C0. Simon does not attend conferences.
###
\end{promptbox}

\begin{promptbox}{ProverQA Hybrid one-shot assistant message}
{
  "goal_analysis": {
    "main_connective": "negation",
    "true_goal": {
      "reasoning": "The conclusion is a negated atomic statement, so the true goal is to derive that Simon does not attend conferences.",
      "composition": "single",
      "statement": "Simon does not attend conferences.",
      "subgoals": [
        {
          "reasoning": "This is the negated atomic conclusion itself.",
          "id": "T1",
          "role": "main_goal",
          "statement": "Simon does not attend conferences."
        }
      ]
    },
    "false_goal": {
      "reasoning": "To make a negated atomic conclusion false, the corresponding positive atomic statement must be true.",
      "composition": "single",
      "statement": "Simon attends conferences.",
      "subgoals": [
        {
          "reasoning": "This is the positive form of the conclusion.",
          "id": "F1",
          "role": "positive_goal",
          "statement": "Simon attends conferences."
        }
      ]
    }
  },
  "proof_tree": {
    "derivations": [
      {
        "reasoning": "P7 directly states that Simon earns recognition.",
        "id": "D1",
        "claim": "Simon earns recognition.",
        "support": ["P7"],
        "rule": "direct_premise"
      },
      {
        "reasoning": "P8 says that if someone earns recognition in their field, then they have expertise. Since Simon earns recognition, Simon has expertise.",
        "id": "D2",
        "claim": "Simon has expertise.",
        "support": ["P8", "D1"],
        "rule": "modus_ponens"
      },
      {
        "reasoning": "P5 says that if Simon has expertise, then he will publish papers. Since Simon has expertise, Simon publishes papers.",
        "id": "D3",
        "claim": "Simon publishes papers.",
        "support": ["P5", "D2"],
        "rule": "modus_ponens"
      },
      {
        "reasoning": "If Simon publishes papers, then Simon either publishes papers or collaborates with peers.",
        "id": "D4",
        "claim": "Simon publishes papers or collaborates with peers.",
        "support": ["D3"],
        "rule": "disjunction_introduction"
      },
      {
        "reasoning": "P13 says that if Simon publishes papers or collaborates with peers, then he attends conferences. Since Simon publishes papers or collaborates with peers, Simon attends conferences.",
        "id": "D5",
        "claim": "Simon attends conferences.",
        "support": ["P13", "D4"],
        "rule": "modus_ponens"
      }
    ],
    "target_matches": [
      {
        "reasoning": "D5 directly matches the false-goal subgoal.",
        "target": "F1",
        "derivations": ["D5"],
        "match_type": "exact"
      },
      {
        "reasoning": "D5 directly proves the false goal.",
        "target": "F",
        "derivations": ["D5"],
        "match_type": "exact"
      }
    ],
    "verdict": {
      "reasoning": "The premises derive that Simon attends conferences, which is the negation of the conclusion. Therefore the conclusion is false.",
      "label": "False",
      "satisfied_target": "F"
    }
  }
}
\end{promptbox}

\subsection{Retry prompt}

The same retry-message template is used when a structured response cannot
be parsed or otherwise cannot be used by the pipeline; the previous
output and the parser or validation error are substituted into the two
placeholders.

\begin{promptbox}{Retry prompt}
Your previous response could not be used.

Previous output:
{previous_output}

Error:
{error_message}

Return corrected JSON only.
\end{promptbox}

\end{document}